\documentclass{article} 
\usepackage{unigs_preprint,times}
\usepackage{amsmath,amsfonts,bm}

\def\eqref#1{equation~\ref{#1}}

\def\1{\bm{1}}

\DeclareMathAlphabet{\mathsfit}{\encodingdefault}{\sfdefault}{m}{sl}
\SetMathAlphabet{\mathsfit}{bold}{\encodingdefault}{\sfdefault}{bx}{n}

\usepackage{hyperref}
\usepackage{url}
\usepackage{booktabs}
\usepackage{array}
\usepackage{multirow}
\usepackage{graphicx}
\usepackage{wrapfig}
\makeatletter
\newcommand{\finishwrap}{%
  \par
  \ifnum\c@WF@wrappedlines>1
    \vspace{\dimexpr\c@WF@wrappedlines\baselineskip-\baselineskip\relax}%
  \fi
  \WFclear}
\makeatother
\usepackage{capt-of}
\usepackage{needspace}
\usepackage{placeins}
\newcommand{\tablefont}{\fontsize{9}{10}\selectfont}
\usepackage[table]{xcolor}
\definecolor{best}{RGB}{248,191,193}    
\definecolor{second}{RGB}{252,216,177}  
\definecolor{third}{RGB}{255,249,177}   

\title{UniGS-SLAM: Monocular Gaussian SLAM with Enhanced Geometry and Open-Vocabulary Semantics}

\author{
  Xingyuan Ding\textsuperscript{1,*} \quad
  Sicheng Yu\textsuperscript{2,*} \quad
  Kaiyong Zhao\textsuperscript{3} \quad
  Hao Wang\textsuperscript{2,\textdagger} \\
  {\normalfont\textsuperscript{1}Shanghai Jiao Tong University} \\
  {\normalfont\textsuperscript{2}The Hong Kong University of Science and Technology (Guangzhou)} \\
  {\normalfont\textsuperscript{3}XGRIDS}
}

\begin{document}

\maketitle

\begingroup
\renewcommand{\thefootnote}{}
\footnotetext[0]{\textsuperscript{*}Equal contribution.\quad
\textsuperscript{\textdagger}Corresponding author:
\href{mailto:haowang@hkust-gz.edu.cn}{\texttt{haowang@hkust-gz.edu.cn}}.}
\endgroup

\begin{figure}[!h]
    \centering
    \includegraphics[width=0.88\linewidth]{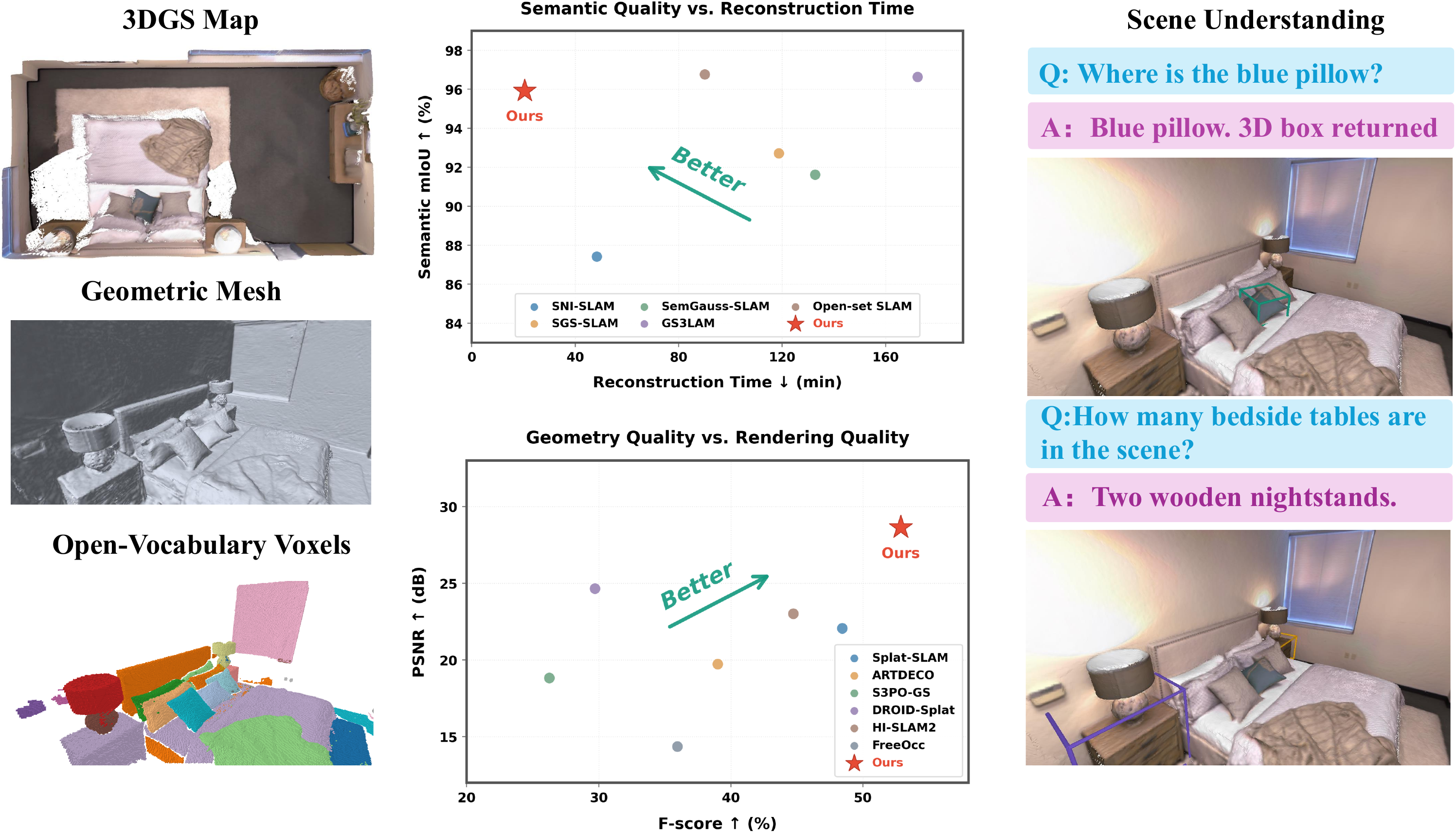}
    \vspace{-5pt}
    \caption{Our RGB-only framework reconstructs detailed geometry and open-vocabulary
  semantics (left), achieves strong quality and efficiency (middle), and supports 3D scene understanding (right).}
    \label{fig:teaser}
\end{figure}

\begin{abstract}
Monocular semantic SLAM aims to reconstruct scene geometry and build semantic maps from RGB observations. Existing Gaussian-based methods often attach semantic labels or features to evolving Gaussian primitives, making semantic fusion difficult to maintain and increasing reconstruction overhead. Hybrid Gaussian--voxel representations provide more stable semantic support, but often rely on RGB-D input or known poses and do not revise historical semantic evidence after geometry updates. We present \textbf{UniGS-SLAM}, an RGB-only framework that decouples open-vocabulary semantics from Gaussian reconstruction while preserving bidirectional interaction. Foundation priors enhance monocular depth and pose estimation, while semantic observations are fused into an independent sparse voxel map with observation provenance, enabling historical contributions to be revised as reconstruction evolves. The fused semantics further guide Gaussian filtering and local surface refinement. Experiments on Replica, ScanNet, and KITTI-360 show leading tracking, rendering, and geometric reconstruction performance among representative RGB-only methods, together with strong open-set semantic mapping on Replica and ScanNet. We also achieve a 2.37$\times$ reconstruction speedup over the fastest compared semantic SLAM baseline. Project page: https://3dagentworld.github.io/UniGS-SLAM/.
\end{abstract}

\section{INTRODUCTION}

Recent advances in 3D Gaussian Splatting (3DGS)~\citep{zhao2026rasterization} have enabled efficient and high-fidelity SLAM systems with explicit scene representations and real-time rendering capabilities~\citep{matsuki2024gaussian,sandstrom2025splat}. Beyond geometric reconstruction, embodied AI applications require semantic maps that associate open-vocabulary concepts with objects and their spatial locations, enabling tasks such as language-guided navigation, object search, and interaction~\citep{yang2025opengsslam,deng2025omnimap}. Building such semantic-aware maps from a single RGB camera is particularly challenging, as semantic understanding relies on accurate geometry while the underlying scene representation continuously evolves during SLAM optimization~\citep{yoo2025openmonogs}.

Early semantic Gaussian SLAM methods attach semantic labels or feature representations directly to Gaussian primitives, enabling unified reconstruction and semantic understanding~\citep{li2024sgsslam,zhu2025semgauss}. More recent approaches incorporate foundation-model semantics to support open-vocabulary scene understanding through explicit Gaussian labels or semantic feature representations~\citep{yang2025opengsslam,yan2026opensetslam}. However, tightly coupling semantics with Gaussian primitives introduces two fundamental limitations. First, Gaussian primitives are continuously optimized, split, and pruned during reconstruction, making their spatial support unstable and complicating consistent semantic fusion for the same physical surfaces. Second, semantic attributes become part of the Gaussian optimization and rendering pipeline, introducing additional computational overhead and limiting the efficiency of online semantic reconstruction.

A natural solution is to decouple semantic representation from Gaussian primitives using more stable spatial structures. Hybrid Gaussian--voxel approaches maintain semantics in structured voxel spaces, providing stable units for multi-view semantic fusion~\citep{yang2025opengsfusion,deng2025omnimap}. However, existing systems typically rely on RGB-D observations and accurate camera poses, which limits their applicability to monocular settings. Moreover, incremental semantic fusion does not explicitly revise historical semantic evidence when backend SLAM optimization updates previous poses or geometry. As a result, semantic observations fused under outdated geometry may remain inconsistent even after the underlying reconstruction is corrected.

To address these challenges, we propose UniGS-SLAM, which decouples semantic representation from Gaussian optimization while preserving bidirectional interaction between semantics and geometry. Gaussian primitives remain dedicated to geometric reconstruction, whereas open-vocabulary semantics are fused in an independent sparse voxel map with observation-level provenance, allowing semantic evidence to be revised after geometric updates. Meanwhile, the fused semantic information is further used to refine Gaussian geometry through semantic-guided filtering and surface refinement.

Specifically, we first enhance monocular Gaussian SLAM with foundation-model geometry priors through geometric context selection, anchor alignment, confidence-aware depth fusion, and long-horizon pose guidance, providing reliable geometry for semantic lifting. We then maintain open-vocabulary semantics in a spatially decoupled sparse voxel representation, where observation-level contributions can be updated when SLAM corrects historical poses or depths. Finally, the semantic map feeds back into reconstruction by guiding floater removal and local surface refinement, allowing semantic understanding to improve Gaussian geometry.

Extensive experiments demonstrate that our RGB-only framework achieves strong performance in geometric reconstruction, rendering quality, and open-vocabulary semantic mapping. We further show that semantic decoupling enables efficient online reconstruction, while semantic-guided refinement improves the quality of the reconstructed Gaussian scene.

Our main contributions are summarized as follows:

\textbf{Geometry-Enhanced Monocular Semantic Gaussian SLAM.}
We enhance monocular Gaussian SLAM with foundation-model geometry priors and confidence-aware fusion, providing reliable geometric support for reconstruction and semantic lifting.

\textbf{Spatially Decoupled Open-Vocabulary Semantic Mapping.}
We maintain open-vocabulary semantics in an independent sparse voxel representation with observation-level provenance, enabling semantic evidence revision after geometric updates.

\textbf{Semantic-Guided Gaussian Geometry Refinement.}
We use fused semantic evidence to guide floater filtering and local surface refinement, allowing semantic understanding to directly improve Gaussian reconstruction.

\section{RELATED WORK}
\label{gen_inst}

\textbf{Gaussian Splatting SLAM.}
3D Gaussian Splatting~\citep{kerbl2023gaussians,zhao2026mmgs} enables efficient rendering
and high-fidelity mapping in dense SLAM~\citep{matsuki2024gaussian,keetha2024splatam,yugay2023gaussianslam,huang2024photoslam,cheng2025reggs}. 
Existing systems either track cameras through Gaussian rendering or combine
separate geometric tracking with Gaussian mapping. DROID-Splat~\citep{homeyer2024droidsplat}
extends DROID-SLAM~\citep{teed2021droid} with Gaussian reconstruction, while
HI-SLAM2~\citep{zhang2025hi} advances HI-SLAM~\citep{zhang2023hi} with Gaussian mapping
and adaptive depth-prior alignment. Splat-SLAM~\citep{sandstrom2025splat} further maintains
map consistency under global pose and depth updates. Rendering-based tracking
can struggle under weak geometric constraints, while in RGB-only settings,
monocular geometric ambiguity can lead to
inaccurate pose and depth estimates that subsequently propagate into
Gaussian mapping.

Recent advances use learned geometric priors to strengthen reconstruction~\citep{cheng2025unposed,yu2025rgb}.
MASt3R-SLAM~\citep{murai2025mast3r} demonstrates foundation-based monocular tracking,
and ARTDECO~\citep{li2025artdeco} integrates foundation predictions into structured
Gaussian reconstruction. S3PO-GS~\citep{cheng2025outdoor} addresses outdoor scale
inconsistency through pointmap alignment, while
KiloGS-SLAM~\citep{kilogsslam2026} targets long-term tracking
scalability. Despite this progress, integrating foundation priors requires
reconciling their scale across prediction windows and rejecting
unreliable geometry.

Our method integrates foundation priors into monocular Gaussian SLAM through
cross-window alignment, confidence-aware depth fusion, and long-horizon
pose guidance.

\textbf{Semantic Gaussian Mapping and SLAM.}
Semantic SLAM has explored neural implicit representations in
SNI-SLAM~\citep{zhu2024snislam} and multi-view feature consensus for
open-vocabulary understanding in RADIO-ViPE~\citep{nasser2026radiovipe}.
For Gaussian-based semantic reconstruction,
OpenGS-Fusion~\citep{yang2025opengsfusion} and OmniMap~\citep{deng2025omnimap}
combine Gaussian appearance modeling with volumetric semantic fusion, but
require RGB-D input and supplied poses.

Semantic Gaussian SLAM brings together online localization, high-fidelity
rendering, and scene understanding. Early methods attach semantic labels or
features to Gaussians~\citep{li2024sgsslam,zhu2025semgauss,li2024gs3lam}, but
are restricted to predefined categories. OpenGS-SLAM~\citep{yang2025opengsslam},
Open-Set Semantic Gaussian SLAM~\citep{yan2026opensetslam}, and
FeatureSLAM~\citep{thirgood2026featureslam} incorporate foundation-model
semantics to support open-set scene understanding, with Open-Set Semantic
Gaussian SLAM further reducing storage through a shared expandable feature
pool. However, these RGB-D systems tie semantics to evolving Gaussian primitives
through labels, features, or indexing vectors; feature-based approaches
also incur additional semantic optimization and rendering costs.
This coupling complicates semantic maintenance, and cross-view agreement
alone cannot correct historical evidence fused under outdated poses or depths.

Our RGB-only framework fuses open-vocabulary semantics in an independent
voxel map and revises historical contributions after pose or depth updates.

\section{METHODOLOGY}
\label{headings}

\begin{figure}[t]
    \centering
    \includegraphics[width=0.94\linewidth]{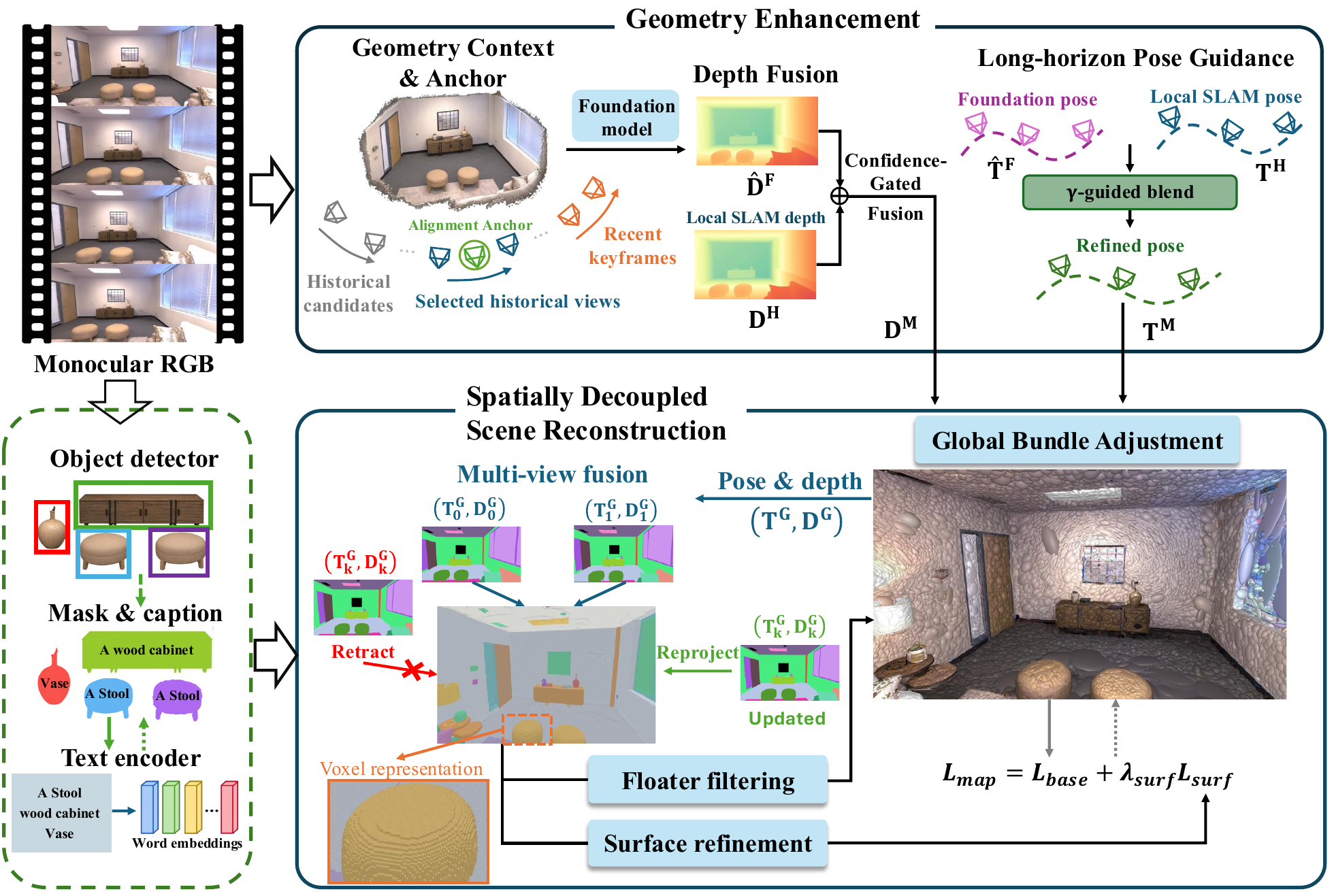}
    \caption{Overview of UniGS-SLAM. Foundation geometry priors enhance monocular reconstruction. An independent voxel map fuses semantic observations and revises historical evidence after geometry updates. Semantic feedback guides Gaussian floater filtering and surface refinement.}
    \label{fig:pipeline}
\end{figure}

As illustrated in Fig.~\ref{fig:pipeline}, our framework takes a calibrated
monocular RGB sequence as input and maintains a Gaussian map for geometry
and appearance together with an independent sparse voxel map for open-vocabulary
semantics. We first use foundation priors to improve monocular tracking
and Gaussian reconstruction (Sec.~\ref{sec:geometry}). Based on this
geometry, we fuse and revise semantic observations in the voxel map
(Sec.~\ref{sec:semantic_voxel}). The resulting semantic map then provides
feedback to refine Gaussian geometry (Sec.~\ref{sec:gaussian_mapping}). Additional method and implementation details are provided in Appendix~\ref{app:method_details}.

\subsection{Geometry-Enhanced Monocular Gaussian SLAM}
  \label{sec:geometry}

  Monocular SLAM methods~\citep{teed2021droid} estimate camera poses and scene
  geometry from visual observations but remain subject to monocular
  geometric ambiguity. Recent geometry foundation models~\citep{Wang_2025_CVPR,depthanything3} provide complementary geometry
  over broader visual contexts, although their predictions may exhibit
  scale and pose inconsistencies across windows. We integrate these
  priors into monocular SLAM through geometric context selection,
  anchor alignment, confidence-aware depth fusion, and long-horizon
  pose guidance.

    \paragraph{Geometric context and anchor alignment.}
  We construct each prediction window from recent tracking keyframes and
  reliable historical views, increasing geometric diversity while
  preserving overlap. For candidate views $i$ and $j$, we measure their
  geometric separation as
  \begin{equation}
  d(i,j)
  =
  \lambda_t\|\mathbf t_i^F-\mathbf t_j^F\|_2
  +
  \frac{1}{\pi}\cos^{-1}\!\left(
  \frac{\operatorname{tr}((R_i^F)^\top R_j^F)-1}{2}
  \right),
  \label{eq:context_distance}
  \end{equation}
  where $(R_i^F,\mathbf t_i^F)$ denotes a previously estimated foundation
  pose and $\lambda_t$ balances translation and rotation.
  We favor confident historical views that complement the recent
  keyframes under this measure. For cross-window consistency, we select
  an alignment anchor using its historical translation and rotation
  residuals, and align the new prediction through the anchor and
  overlapping observations. This provides longer-range context within
  a bounded inference window.

  \paragraph{Confidence-aware depth fusion.}
  Let $D_t^S$ denote local SLAM depth and $\widetilde D_t^F$ the
  foundation depth aligned to the current SLAM scale over reliable
  overlapping pixels. We determine the foundation contribution from
  prediction confidence and agreement with local geometry. For valid
  pixels in the blending regime, the weight is
  \begin{equation}
  \alpha_t(\mathbf u)
  =
  \bar C_t^F(\mathbf u)
  \exp\!\left(
  -\frac{
  |\log \widetilde D_t^F(\mathbf u)-\log D_t^S(\mathbf u)|
  }{\tau_e}
  \right),
  \label{eq:depth_gate}
  \end{equation}
  where $\bar C_t^F$ is the confidence rescaled to the blending interval
  and $\tau_e$ controls disagreement sensitivity.
  Unreliable or inconsistent predictions are rejected, while
  high-confidence consistent predictions receive full weight.
  We blend the aligned depths as
  \begin{equation}
  \bar D_t(\mathbf u)
  =
  (1-\alpha_t(\mathbf u))D_t^S(\mathbf u)
  +
  \alpha_t(\mathbf u)\widetilde D_t^F(\mathbf u).
  \label{eq:depth_fusion}
  \end{equation}
  A smooth residual scale correction, modulated by the same confidence
  gate, compensates for low-frequency alignment bias to obtain the
  mapping depth $D_t^M$.
  We initialize Gaussian primitives by back-projecting $D_t^M$ with the
  estimated camera poses, and optimize their geometry and appearance
  through differentiable rendering with photometric and geometric
  supervision~\citep{matsuki2024gaussian}.
  
  \paragraph{Long-horizon pose guidance.}
  The historical context also provides trajectory proposals beyond the
  local tracking window. We align these proposals to the SLAM coordinate
  system using a $\mathrm{Sim}(3)$ transformation and screen them for
  geometric consistency. Accepted proposals are blended with the current
  poses to initialize global visual bundle adjustment, with a coefficient
  $\gamma$ controlling the guidance strength. Bundle adjustment then
  refines poses and depths under the visual correspondence constraints,
  allowing observation-driven optimization to correct residual prior
  errors. The updated geometry is propagated to Gaussian reconstruction.

\subsection{Spatially Decoupled Open-Vocabulary Semantic Mapping}
  \label{sec:semantic_voxel}
We maintain persistent open-vocabulary semantics in an independent sparse
  voxel map, retaining observation-level contributions to support revision
  as the reconstructed geometry evolves.

  \paragraph{Observation-based voxel fusion.}
  For each semantic keyframe, a region-language frontend extracts region
  masks, normalized language embeddings, and confidence
  scores~\citep{deng2025omnimap}.
  Using the current Gaussian-rendered depth and corresponding camera pose,
  we lift each region observation $o$ into the voxel map.
  Its contribution to voxel $v$ is denoted by
  $\Delta\mathbf z_{ov}=w_{ov}\mathbf e_o$, where $\mathbf e_o$ is the
  language embedding and $w_{ov}$ is its fusion weight, which is zero outside
  the observation's voxel support.
  Each voxel aggregates these contributions into a normalized semantic feature, enabling open-vocabulary retrieval through similarity comparison with text-query embeddings.
  We also associate observations across views using spatial overlap
  and language similarity to maintain entity identities.
  Alongside the fused statistics, we retain the original region
  observations and their individual voxel contributions to support subsequent geometry-aware semantic revisions.

  \paragraph{Revision under geometry updates.}
  Changes to camera poses or Gaussian geometry can invalidate the spatial
  support of previously fused observations.
  For each affected observation, we first retract its stored contribution,
  then reproject the cached region observation using the updated depth
  and pose and insert its new contribution:
  \begin{equation}
  \mathbf z_v=\sum_o\Delta\mathbf z_{ov},
  \qquad
  \mathbf z_v'
  =
  \mathbf z_v
  -\Delta\mathbf z_{ov}^{\mathrm{old}}
  +\Delta\mathbf z_{ov}^{\mathrm{new}}.
  \label{eq:semantic_revision}
  \end{equation}
  The accumulated weights are updated analogously, and the revised
  semantic feature is obtained by normalizing $\mathbf z_v'$.
  The cached masks and language embeddings are reused, so geometry
  revision requires no semantic re-encoding; unaffected observations
  require no re-fusion.

\subsection{Semantic-Guided Gaussian Geometry Refinement}
  \label{sec:gaussian_mapping}

  Voxel semantics guide the selection of floater candidates and compatible
  surface neighborhoods for Gaussian refinement.
  Geometric and photometric evidence determines the corrections.

  \paragraph{Verified floater filtering.}
  Weak or missing semantic support identifies potential floaters in
  spatially sparse Gaussian neighborhoods, but is insufficient to justify
  removal. We verify these candidates using counter-evidence from multiple
  distinct views: either repeated free-space violations relative to
  reliable predicted mapping depths $D_t^M$ from
  Sec.~\ref{sec:geometry}, or photometric evidence that
  reducing their opacity improves image reconstruction.
  Occluded Gaussians are not treated as free-space violations, and
  unobserved or semantically uncovered regions remain unknown.
  For verified candidates, we attenuate opacity and subsequently prune
  only those with sufficiently low opacity and renewed multi-view
  counter-evidence. This filtering is implemented as a map-management procedure without introducing an additional semantic loss term.

\begin{figure}[!t]
    \centering
    \includegraphics[width=\linewidth,keepaspectratio]{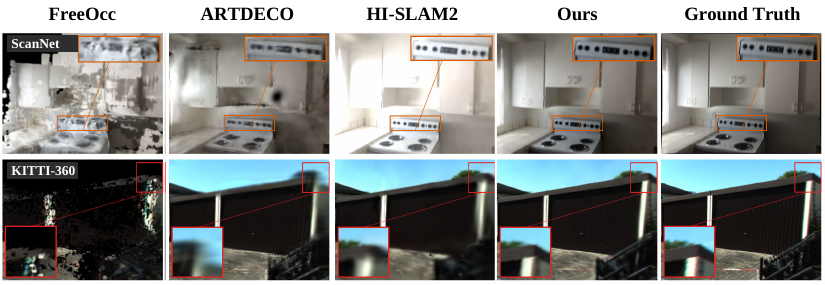}
    \caption{RGB rendering comparisons on ScanNet and KITTI-360.
Our method recovers finer scene details and achieves higher rendering quality.}
    \label{fig:rendering_comparison}
    
\par\medskip
\begingroup
\centering

\captionof{table}{Tracking and rendering comparisons on Replica and ScanNet.
The \colorbox{best}{best}, \colorbox{second}{second-best}, and
\colorbox{third}{third-best} results among RGB-only methods are highlighted,
respectively.}
\label{tab:tracking_rendering_indoor}
\par\vspace{6pt}
\setlength{\tabcolsep}{4.2pt}
\renewcommand{\arraystretch}{0.88}
\resizebox{\textwidth}{!}{%
\begin{tabular}{lc|cccc|cccc}
\toprule
\multirow{2}{*}{\textbf{Method}} &
\multirow{2}{*}{\textbf{Input}} &
\multicolumn{4}{c|}{\textbf{Replica}} &
\multicolumn{4}{c}{\textbf{ScanNet}} \\
\cmidrule(lr){3-6}
\cmidrule(lr){7-10}

& &
\textbf{ATE $\downarrow$}
& \textbf{PSNR $\uparrow$}
& \textbf{SSIM $\uparrow$}
& \textbf{LPIPS $\downarrow$}
& \textbf{ATE $\downarrow$}
& \textbf{PSNR $\uparrow$}
& \textbf{SSIM $\uparrow$}
& \textbf{LPIPS $\downarrow$} \\
\midrule


Open-set SLAM
& RGB-D
& 0.230
& 37.95
& 0.989
& 0.053
& 7.54
& --
& --
& -- \\

GS3LAM
& RGB-D
& 0.370
& 36.26
& 0.989
& 0.052
& 11.24
& 22.86
& 0.868
& 0.222 \\

SemGauss-SLAM
& RGB-D
& 0.380
& 34.83
& 0.978
& 0.069
& 9.46
& 23.75
& 0.821
& 0.241 \\

\midrule


Splat-SLAM
& RGB
& 0.446
& 34.15
& 0.945
& 0.074
& 7.75
& 22.07
& 0.792
& \cellcolor{best}0.230 \\

ARTDECO
& RGB
& 1.717
& 30.33
& 0.910
& 0.171
& 7.63
& 19.73
& 0.784
& 0.357 \\

S3PO-GS
& RGB
& 2.291
& 30.05
& 0.895
& 0.203
& 35.10
& 18.82
& 0.772
& 0.571 \\

DROID-Splat
& RGB
& 0.408
& 34.56
& 0.944
& 0.073
& 23.82
& \cellcolor{second}24.65
& \cellcolor{third}0.800
& 0.295 \\

HI-SLAM2
& RGB
& \cellcolor{second}0.251
& \cellcolor{second}37.72
& \cellcolor{third}0.964
& \cellcolor{second}0.034
& \cellcolor{second}7.21
& \cellcolor{third}23.02
& \cellcolor{second}0.818
& \cellcolor{third}0.269 \\

FreeOcc
& RGB
& 0.941
& 19.03
& 0.693
& 0.280
& \cellcolor{third}7.44
& 14.36
& 0.446
& 0.563 \\

Hier-SLAM++
& RGB
& \cellcolor{third}0.310
& \cellcolor{third}35.68
& \cellcolor{best}0.981
& \cellcolor{third}0.066
& 11.59
& --
& --
& -- \\

\midrule

\textbf{Ours}
& RGB
& \cellcolor{best}0.247
& \cellcolor{best}38.65
& \cellcolor{second}0.975
& \cellcolor{best}0.026
& \cellcolor{best}7.15
& \cellcolor{best}28.65
& \cellcolor{best}0.859
& \cellcolor{second}0.239 \\

\bottomrule
\end{tabular}}
\par
\normalsize
\endgroup
\end{figure}

\paragraph{Semantic-compatible surface refinement.}
  For each Gaussian, we select local supporting voxels based on spatial proximity, language-feature similarity, and entity-assignment compatibility.
  We recover surface samples from predicted mapping depths $D_t^M$
  across views, retaining samples that satisfy confidence, semantic
  compatibility, and geometric consistency checks.
  Using the retained samples within the selected voxel support, we
  estimate a local center $\bar{\mathbf x}_j$ and unit normal
  $\mathbf n_j$ by weighted PCA.
  Compared with purely spatial neighborhood selection, incorporating semantic and entity compatibility helps prevent the mixing of nearby but distinct object surfaces. These cues guide sample grouping, while surface positions and orientations are estimated from geometry.
  We retain only neighborhoods with sufficient multi-view support,
  geometric consistency, and local planarity.

  Let $\boldsymbol\mu_j$ and $\Sigma_j$ denote the center and covariance
  of Gaussian $j$. We first encourage its center to approach the
  estimated surface through a robust point-to-plane constraint:
  \begin{equation}
  \mathcal L_P
  =
  \left\langle
  \rho\!\left(
  \frac{
  \mathbf n_j^\top(\boldsymbol\mu_j-\bar{\mathbf x}_j)
  }{h}
  \right)
  \right\rangle_{w_j},
  \label{eq:semantic_surface_position}
  \end{equation}
  where $h$ is the voxel size, $\rho$ is a smooth-$L_1$ penalty, and
  $\langle\cdot\rangle_{w_j}$ denotes averaging weighted by surface
  reliability.
  We further suppress excessive Gaussian extent along the surface normal:
  \begin{equation}
\begin{aligned}
\mathcal{L}_H
= \Big\langle
\log\Big(
1+\max\Big(
&\frac{\mathbf{n}_j^\top\Sigma_j\mathbf{n}_j}{h^2}
-\tau_H^2,
&0
\Big)
\Big)
\Big\rangle_{w_j},
\end{aligned}
\label{eq:semantic_surface_thickness}
\end{equation}
  where $\tau_H$ specifies the permitted
  normal-direction thickness relative to the voxel size.
  The position term adjusts Gaussian centers, while the thickness term
  acts on their scales and orientations without directly penalizing
  tangential extent.

\paragraph{Optimization objective.}
  We define the base reconstruction and surface losses as
  \begin{equation}
  \mathcal L_{\mathrm{base}}
  =
  \lambda_{\mathrm{rgb}}\mathcal L_{\mathrm{rgb}}
  +\lambda_{\mathrm{depth}}\mathcal L_{\mathrm{depth}}
  +\lambda_{\mathrm{normal}}\mathcal L_{\mathrm{normal}},
  \qquad
  \mathcal L_{\mathrm{surf}}
  =
  \mathcal L_P+\lambda_H\mathcal L_H.
  \label{eq:base_surface_losses}
  \end{equation}
  The combined mapping objective is
  \begin{equation}
  \mathcal L_{\mathrm{map}}
  =
  \mathcal L_{\mathrm{base}}
  +\lambda_{\mathrm{surf}}\mathcal L_{\mathrm{surf}}.
  \label{eq:mapping_objective}
  \end{equation}
  Surface constraints apply only to reliable neighborhoods, with targets
  held fixed during the corresponding optimization.

\setlength{\textfloatsep}{8pt}
\setlength{\floatsep}{8pt}

\begin{figure}[!t]
    \centering
    \includegraphics[width=0.96\linewidth,keepaspectratio]{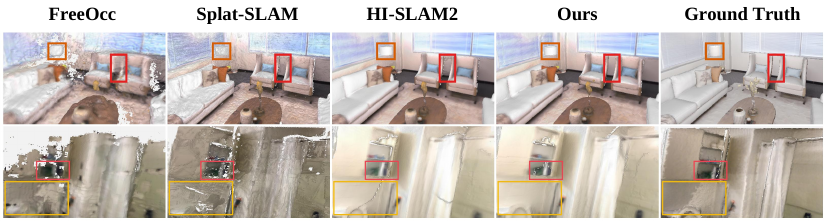}
    \caption{Mesh comparisons on Replica and ScanNet, showing cleaner surfaces and finer details.}
\label{fig:mesh_comparison}
\par\smallskip
\begin{minipage}[t]{\dimexpr0.6\textwidth-0.6\columnsep\relax}
\vspace{0pt}

\centering

\captionof{table}{Geometric reconstruction on Replica and ScanNet.}
\label{tab:geometry_comparison}
\par\vspace{6pt}
\setlength{\tabcolsep}{0.7pt}
\renewcommand{\arraystretch}{0.88}
\tablefont
\begin{tabular*}{\linewidth}{@{\extracolsep{\fill}}lcccccc}
\toprule
\multirow{2}{*}{\textbf{Method}} &
\multicolumn{3}{c}{\textbf{Replica}} &
\multicolumn{3}{c}{\textbf{ScanNet}} \\
\cmidrule(lr){2-4}
\cmidrule(lr){5-7}

& \shortstack{\textbf{Acc.}$\downarrow$}
& \shortstack{\textbf{L1}$\downarrow$}
& \shortstack{\textbf{F-score}$\uparrow$}
& \shortstack{\textbf{Acc.}$\downarrow$}
& \shortstack{\textbf{L1}$\downarrow$}
& \shortstack{\textbf{F-score}$\uparrow$} \\
\midrule

ARTDECO
& 12.15
& 12.565
& 57.43
& 63.46
& \cellcolor{third}14.467
& 39.01 \\

DROID-Splat
& 7.07
& 5.690
& 66.59
& 39.08
& 19.400
& 29.70 \\

Splat-SLAM
& \cellcolor{third}2.89
& \cellcolor{third}2.468
& \cellcolor{third}82.73
& \cellcolor{third}16.40
& 14.663
& \cellcolor{second}48.42 \\

FreeOcc
& 4.25
& 12.820
& 63.37
& \cellcolor{second}15.23
& 19.716
& 35.94 \\

S3PO-GS
& 173.06
& 11.523
& 20.83
& 22.70
& 42.427
& 26.25 \\

HI-SLAM2
& \cellcolor{second}2.01
& \cellcolor{second}1.528
& \cellcolor{second}86.64
& 19.68
& \cellcolor{second}10.467
& \cellcolor{third}44.73 \\
\midrule
\textbf{Ours}
& \cellcolor{best}1.48
& \cellcolor{best}1.170
& \cellcolor{best}89.84
& \cellcolor{best}9.18
& \cellcolor{best}9.388
& \cellcolor{best}52.86 \\

\bottomrule
\end{tabular*}
\par
\normalsize

\end{minipage}%
\hspace{\columnsep}%
\begin{minipage}[t]{\dimexpr0.4\textwidth-0.4\columnsep\relax}
\vspace{0pt}
\centering

\captionof{table}{Tracking and rendering on KITTI-360.}
\label{tab:kitti360}
\par\vspace{6pt}
\setlength{\tabcolsep}{0.4pt}
\renewcommand{\arraystretch}{1.129}

\setlength{\aboverulesep}{0.25ex}
\setlength{\belowrulesep}{0.25ex}

\tablefont
\resizebox{\linewidth}{!}{%
\begin{tabular}{@{}lcccc@{}}
\toprule
\textbf{Method}
& \shortstack{\textbf{ATE}$\downarrow$}
& \shortstack{\textbf{PSNR}$\uparrow$}
& \shortstack{\textbf{SSIM}$\uparrow$}
& \shortstack{\textbf{LPIPS}$\downarrow$} \\
\midrule

Splat-SLAM
& \cellcolor{second}1.515
& 23.40
& 0.729
& 0.367 \\

ARTDECO
& 11.534
& 22.00
& 0.714
& \cellcolor{second}0.278 \\

S3PO-GS
& 12.162
& 20.65
& 0.668
& 0.407 \\

DROID-Splat
& 2.075
& \cellcolor{second}23.73
& \cellcolor{third}0.752
& \cellcolor{third}0.324 \\

FreeOcc
& 3.823
& 9.36
& 0.326
& 0.533 \\

HI-SLAM2
& \cellcolor{third}1.668
& \cellcolor{third}23.42
& \cellcolor{second}0.754
& 0.339 \\
\midrule
\textbf{Ours}
& \cellcolor{best}1.366
& \cellcolor{best}25.39
& \cellcolor{best}0.800
& \cellcolor{best}0.261 \\

\bottomrule
\end{tabular}}
\par
\normalsize

\end{minipage}
\end{figure}

\section{EXPERIMENT}
\subsection{EXPERIMENT SETUP}
\noindent\textbf{Datasets.}
Experiments are conducted on eight scenes from
Replica~\citep{straub2019replica}, six scenes from
ScanNet~\citep{dai2017scannet}, and four static segments from
KITTI-360~\citep{liao2022kitti360}.
Dataset and scene-selection details are provided in
Appendix~\ref{app:dataset_details}.

\noindent\textbf{Baselines.}
We compare with representative visual SLAM
methods~\citep{sandstrom2025splat,li2025artdeco,cheng2025outdoor,homeyer2024droidsplat,zhang2025hi} and semantic
mapping systems~\citep{zhu2024snislam,li2024sgsslam,zhu2025semgauss,li2025hierslampp,yang2025opengsslam,li2024gs3lam,yan2026opensetslam,jiang2026freeocc,nasser2026radiovipe}.

\noindent\textbf{Implementation Details.}
All experiments are run on a server equipped with a single NVIDIA RTX A6000
GPU. Our system uses Depth Anything 3~\citep{depthanything3} as geometric foundation model.
\subsection{TRACKING AND APPEARANCE MAPPING RESULTS}

We evaluate camera tracking, appearance reconstruction, and geometric
reconstruction. Our RGB-only system achieves strong tracking accuracy
and rendering quality in both indoor and outdoor scenes, together with
substantial improvements in indoor geometric reconstruction.

\noindent\textbf{Metrics.}
We evaluate RGB rendering using PSNR (dB), SSIM~\citep{wang2004ssim},
and LPIPS~\citep{zhang2018lpips}. Camera tracking is measured by the
root mean squared absolute trajectory error (ATE RMSE). Geometric
reconstruction is assessed using accuracy (Acc.), L1 depth error (L1),
and F-score (\%).

\noindent\textbf{Rendering and Camera Pose Estimation.}
Table~\ref{tab:tracking_rendering_indoor} shows that our method achieves
the highest PSNR among all compared methods on both indoor datasets,
exceeding the strongest RGB-only baseline by 4.00\,dB on ScanNet.
On Replica, it surpasses the RGB-D baseline Open-set
SLAM~\citep{yan2026opensetslam} by 0.70\,dB and achieves the lowest LPIPS.
It also achieves the lowest ATE among RGB-only methods on Replica and
tracking accuracy comparable to HI-SLAM2 on ScanNet.
Beyond indoor scenes, our method leads across all four metrics on the
selected KITTI-360 segments (Table~\ref{tab:kitti360}, ATE in meters),
reducing ATE by 9.8\% and improving PSNR by 1.66\,dB over the best baseline
for each metric.

\noindent\textbf{Geometric Reconstruction.}

As shown in Table~\ref{tab:geometry_comparison}, our method achieves
the best results across all three geometric metrics on both Replica
and ScanNet. On Replica, it reduces Acc. and L1 depth error by 26.4\%
and 23.4\%, respectively, relative to HI-SLAM2.
On ScanNet, it reduces Acc. by 39.7\% relative to FreeOcc and improves
F-score by 4.44 percentage points over Splat-SLAM, the strongest
competing RGB-only baselines on the respective metrics.

\noindent\textbf{Qualitative Comparisons of Rendering and Geometry.}
Figure~\ref{fig:rendering_comparison} shows that our method better preserves
appearance details and object boundaries on ScanNet and KITTI-360.
Figure~\ref{fig:mesh_comparison} further demonstrates more regular object
shapes and continuous surfaces on Replica and ScanNet, with fewer holes
and geometric artifacts than the baselines.

\subsection{CLOSED-SET AND OPEN-SET SEMANTIC MAPPING RESULTS}

\begin{figure}[!t]
    \centering
    \includegraphics[width=0.96\linewidth,keepaspectratio]{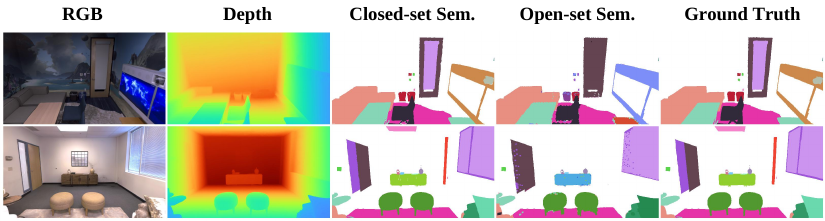}
    \caption{Semantic maps on Replica, showing coherent regions consistent with ground truth.}
\label{fig:semantic_qualitative}

\par\smallskip
\par
\noindent\begin{minipage}{\textwidth}
\setlength{\parskip}{0pt}
\begin{minipage}[t]{\dimexpr0.6\textwidth-0.6\columnsep\relax}
\vspace{0pt}
\centering

\captionof{table}{Closed-set 2D semantic segmentation on Replica.}
\label{tab:semantic_replica}
\par\vspace{6pt}
\setlength{\tabcolsep}{0.6pt}
\renewcommand{\arraystretch}{0.8135}
\tablefont
\begin{tabular*}{\linewidth}{@{\extracolsep{\fill}}lcccccc}
\toprule
\textbf{Method} &
\textbf{Input} &
\textbf{R0} &
\textbf{R1} &
\textbf{R2} &
\textbf{O0} &
\textbf{Avg.} \\
\midrule

SNI-SLAM
& RGB-D
& 88.42
& 87.43
& 86.16
& 87.63
& 87.41 \\

SGS-SLAM
& RGB-D
& 92.95
& 92.91
& 92.10
& 92.90
& 92.72 \\

SemGauss-SLAM
& RGB-D
& 91.63
& 92.72
& 87.51
& 94.60
& 91.62 \\

OpenGS-SLAM
& RGB-D
& 93.24
& 94.11
& 92.79
& 93.22
& 93.34 \\

GS3LAM
& RGB-D
& \cellcolor{second}96.83
& \cellcolor{third}96.68
& \cellcolor{second}96.40
& \cellcolor{second}96.61
& \cellcolor{second}96.63 \\

Open-set SLAM
& RGB-D
& \cellcolor{best}96.93
& \cellcolor{second}96.82
& \cellcolor{best}96.67
& \cellcolor{best}96.62
& \cellcolor{best}96.76 \\

FreeOcc
& RGB
& 68.63
& 75.91
& 60.31
& 67.78
& 68.16 \\

Hier-SLAM++
& RGB
& 91.21
& 90.62
& 89.11
& 90.45
& 90.35 \\
\midrule
\textbf{Ours}
& RGB
& \cellcolor{third}95.39
& \cellcolor{best}96.85
& \cellcolor{third}96.26
& \cellcolor{third}95.19
& \cellcolor{third}95.92 \\

\bottomrule
\end{tabular*}
\par
\normalsize
\end{minipage}%
\hspace{\columnsep}%
\begin{minipage}[t]{\dimexpr0.4\textwidth-0.4\columnsep\relax}
\vspace{0pt}
\centering

\captionof{table}{Open-set 3D semantic segmentation.}
\label{tab:semantic_3d_comparison}
\par\vspace{6pt}
\setlength{\tabcolsep}{0.7pt}

\renewcommand{\arraystretch}{0.9745}
\setlength{\aboverulesep}{0pt}
\setlength{\belowrulesep}{0.25ex}
\tablefont
\begin{tabular*}{\linewidth}{@{\extracolsep{\fill}}lcccc@{}}
\toprule
\textbf{Method} & \textbf{mAcc} & \textbf{fAcc} & \textbf{mIoU} & \textbf{fIoU} \\
\midrule
\multicolumn{5}{c}{\textbf{Replica}} \\
\midrule
FreeOcc & 0.29 & 0.50 & 0.19 & 0.44 \\
RADIO-ViPE & 0.36 & 0.63 & 0.27 & 0.56 \\
\textbf{Ours} & \textbf{0.42} & \textbf{0.73} & \textbf{0.38} & \textbf{0.69} \\
\midrule
\multicolumn{5}{c}{\textbf{ScanNet}} \\
\midrule
FreeOcc & 0.22 & 0.21 & 0.16 & 0.17 \\
RADIO-ViPE & 0.50 & 0.52 & 0.38 & 0.45 \\
\textbf{Ours} & \textbf{0.61} & \textbf{0.66} & \textbf{0.50} & \textbf{0.56} \\
\bottomrule
\end{tabular*}
\par
\normalsize
\end{minipage}
\end{minipage}
\par
\end{figure}

\noindent\textbf{Metrics.}
We measure closed-set 2D segmentation using mean Intersection over Union
(mIoU), reported in \%. For open-set 3D object semantics, we report mean class accuracy
(mAcc), frequency-weighted class accuracy (fAcc), mIoU, and frequency-weighted IoU
(fIoU), following the RADIO-ViPE semantic segmentation protocol.
Higher values indicate better semantic performance.
Table~\ref{tab:semantic_3d_comparison} uses a 0--1 scale, with the best
results in bold. Additional details of the semantic evaluation protocol are provided in
Appendix~\ref{app:evaluation_details}.

\noindent\textbf{Closed-Set Semantic Mapping.}
We evaluate on four Replica scenes with ground-truth semantic supervision
following Open-set SLAM~\citep{yan2026opensetslam}, while our tracking and
reconstruction remain RGB-only.
Table~\ref{tab:semantic_replica} shows that our method outperforms the
compared RGB-only baselines, improving mean mIoU over Hier-SLAM++ by
5.57 percentage points. Despite requiring no sensor depth, it also
surpasses several RGB-D methods and approaches the strongest RGB-D
results: the gaps to GS3LAM and Open-set SLAM are only 0.71 and
0.84 percentage points, respectively.

\noindent\textbf{Open-Set Semantic Mapping.}
We follow the RADIO-ViPE~\citep{nasser2026radiovipe} protocol on Replica
and ScanNet, comparing our RGB-only system with FreeOcc and RADIO-ViPE.
Table~\ref{tab:semantic_3d_comparison} evaluates the semantic quality of
the reconstructed 3D maps. Our method outperforms both FreeOcc and
RADIO-ViPE across all four metrics on both datasets, improving mIoU over
the stronger baseline, RADIO-ViPE, by 11 percentage points on Replica
and 12 percentage points on ScanNet.

\noindent\textbf{Qualitative Semantic Results.}
Figure~\ref{fig:semantic_qualitative} shows our closed-set and open-set
semantic mapping results on Replica. Both recover distinct object regions and preserve their spatial structure.

\subsection{ABLATION STUDY}

We ablate three key components: geometry enhancement, historical semantic
revision, and semantic-guided refinement. We also compare running
time against semantic SLAM baselines.

\FloatBarrier
\begin{wraptable}[10]{r}{\dimexpr0.57\textwidth-0.57\columnsep\relax}
\vspace{\dimexpr-2pt-\intextsep\relax}
\centering

\caption{Geometry enhancement ablation.}
\label{tab:ablation_geometry}
\par\vspace{6pt}
\setlength{\tabcolsep}{0.5pt}
\renewcommand{\arraystretch}{1.08}
\tablefont
\begin{tabular*}{\linewidth}{@{\extracolsep{\fill}}lrrr@{}}
\toprule
\shortstack[l]{Method\\\phantom{[cm] $\downarrow$}} & \shortstack{PSNR\\{[dB]} $\uparrow$} & \shortstack{L1 depth\\{[cm]} $\downarrow$} & \shortstack{ATE RMSE\\{[cm]} $\downarrow$} \\
\midrule
Foundation Prior (DA3) & 33.68 & 1.964 & 2.010 \\
\addlinespace[2pt]
+ Context \& Anchor Alignment & 34.79 & 1.310 & 0.713 \\
\addlinespace[2pt]
+ Depth Fusion & 40.34 & 1.083 & 0.623 \\
\addlinespace[2pt]
+ Long-Horizon Guidance & \textbf{42.61} & \textbf{1.023} & \textbf{0.207} \\
\bottomrule
\end{tabular*}
\par
\normalsize
\end{wraptable}
\noindent\textbf{Geometry Enhancement.}
Table~\ref{tab:ablation_geometry} adds geometry enhancement components to
the DA3 baseline one at a time on Replica office0. Context selection and
anchor alignment improve pose and depth accuracy, while confidence-aware
depth fusion further improves reconstruction quality. Long-horizon pose
guidance provides an additional reduction in trajectory error. Each
component improves all three metrics, demonstrating their complementary
contributions to tracking and reconstruction.
\finishwrap

  \FloatBarrier
  \begin{wraptable}{r}{0.4\textwidth}
  \centering
  \caption{Semantic revision ablation.}
  \label{tab:ablation_semantic_revision}
\par\vspace{6pt}
  \setlength{\tabcolsep}{1pt}
  \renewcommand{\arraystretch}{0.88}
  \tablefont
  \begin{tabular*}{\linewidth}{@{\extracolsep{\fill}}l|cc@{}}
  \toprule
  Method
  & \shortstack{mAcc$\uparrow$}
  & \shortstack{mIoU$\uparrow$} \\
  \midrule
  w/o Historical Revision
  & 55.85
  & 45.43 \\
  Reprojection w/o Retraction
  & 56.90
  & 46.25 \\
  Ours
  & \textbf{60.80}
  & \textbf{49.86} \\
  \bottomrule
  \end{tabular*}
  \par
  \normalsize
  \end{wraptable}

  \noindent\textbf{Historical Semantic Revision.}
  Table~\ref{tab:ablation_semantic_revision} compares three revision
  strategies averaged over all ScanNet scenes.
  Reprojection without retraction yields only modest gains, because the
old semantic contributions remain in the map after the updated
observations are inserted.
  Our retract-and-reintegrate strategy (Eq.~\ref{eq:semantic_revision})
  achieves the best performance, improving mAcc and mIoU by
  4.95 and 4.43 percentage points over no revision, respectively.
  This demonstrates the importance of explicitly removing outdated
  semantic evidence when the underlying geometry is updated.

\noindent\begin{minipage}[t]{\dimexpr0.6\textwidth-6pt\relax}
\vspace{0pt}
\noindent\textbf{Semantic-Guided Geometry Refinement.}
Table~\ref{tab:ablation_semantic_geometry} evaluates semantic-guided
geometry refinement on ScanNet scene0207. Verified floater filtering
reduces surface error and improves F-score, while adding
semantic-compatible surface refinement yields further gains.
These results show that the decoupled semantic map provides effective
feedback for improving Gaussian geometry.
\end{minipage}%
\hspace{6pt}%
\begin{minipage}[t]{0.4\textwidth}
\vspace{0pt}
\raggedright

\captionof{table}{Semantic-guided refinement.}
\label{tab:ablation_semantic_geometry}
\par\vspace{6pt}
\setlength{\tabcolsep}{1pt}
\renewcommand{\arraystretch}{0.88}
\tablefont
\begin{tabular*}{\linewidth}{@{\extracolsep{\fill}}l|cc@{}}
\toprule
Method
& \shortstack{Acc.\\{[cm]} $\downarrow$}
& \shortstack{F-score\\{[\%]} $\uparrow$} \\
\midrule

w/o both
& 13.76
& 48.86 \\

+ Floater Filtering
& 10.08
& 50.28 \\

+ Surface Refinement
& \textbf{9.79}
& \textbf{50.97} \\

\bottomrule
\end{tabular*}
\par
\normalsize
\end{minipage}
\par\smallskip

\noindent\begin{minipage}[t]{\dimexpr0.6\textwidth-6pt\relax}
\vspace{0pt}
\noindent\textbf{Computational Efficiency.}
Table~\ref{tab:replica_total_time} reports reconstruction time on Replica.
Our method takes 20.40 minutes, achieving a $2.37\times$ speedup over
the fastest compared baseline, SNI-SLAM. This efficiency is consistent
with our decoupled design, which fuses language features in voxels without
optimizing or rendering them as Gaussian attributes. Detailed timing
protocols and runtime calculations are provided in
Appendix~\ref{app:evaluation_details}.
\end{minipage}%
\hspace{6pt}%
\begin{minipage}[t]{0.4\textwidth}
\vspace{0pt}
\raggedright

\captionof{table}{Reconstruction time.}
\label{tab:replica_total_time}
\par\vspace{6pt}
\setlength{\tabcolsep}{1pt}
\renewcommand{\arraystretch}{0.88}
\tablefont
\begin{tabular*}{\linewidth}{@{\extracolsep{\fill}}lc@{}}
\toprule
\textbf{Method} & \textbf{Time (min) $\downarrow$} \\
\midrule

SNI-SLAM
& 48.32 \\

SGS-SLAM
& 118.70 \\

SemGauss-SLAM
& 132.79 \\

GS3LAM
& 172.40 \\

\shortstack[l]{Open-set SLAM}
& 90.10 \\
\midrule
\textbf{Ours}
& \textbf{20.40} \\

\bottomrule
\end{tabular*}
\par
\normalsize
\end{minipage}

\Needspace{6\baselineskip}
\section{Conclusion}

This work shows that geometry and semantics in monocular semantic SLAM can be decoupled in representation while remaining connected in optimization. UniGS-SLAM maintains geometry in a Gaussian map and open-vocabulary semantics in an independent voxel map, enabling semantic evidence to follow geometric updates without entering Gaussian optimization and rendering. The semantic map further feeds back to improve Gaussian reconstruction through targeted filtering and surface refinement. Our results demonstrate that this design supports accurate reconstruction, strong semantic mapping, and efficient RGB-only SLAM. More broadly, separating geometric and semantic state while preserving controlled interaction offers a useful design principle for semantic SLAM with evolving scene representations.
\subsection*{AI use statement}
We used large language models to aid manuscript writing and polish the
text, including checking spelling and grammar. The authors take full
responsibility for the final content, including all text, claims, and
conclusions.

\setlength{\textfloatsep}{12pt}
\setlength{\floatsep}{12pt}
\setlength{\intextsep}{12pt}
\setlength{\abovecaptionskip}{10pt}
\bibliography{reference}
\bibliographystyle{reference}

\clearpage
\appendix
\section{Appendix}

\finishwrap
\Needspace{32\baselineskip}
\subsection{APPLICATIONS}
\label{sec:applications}
\begin{wrapfigure}{r}{0.48\textwidth}
    \vspace{-\intextsep}
    \centering
    \includegraphics[width=\linewidth,pagebox=cropbox]{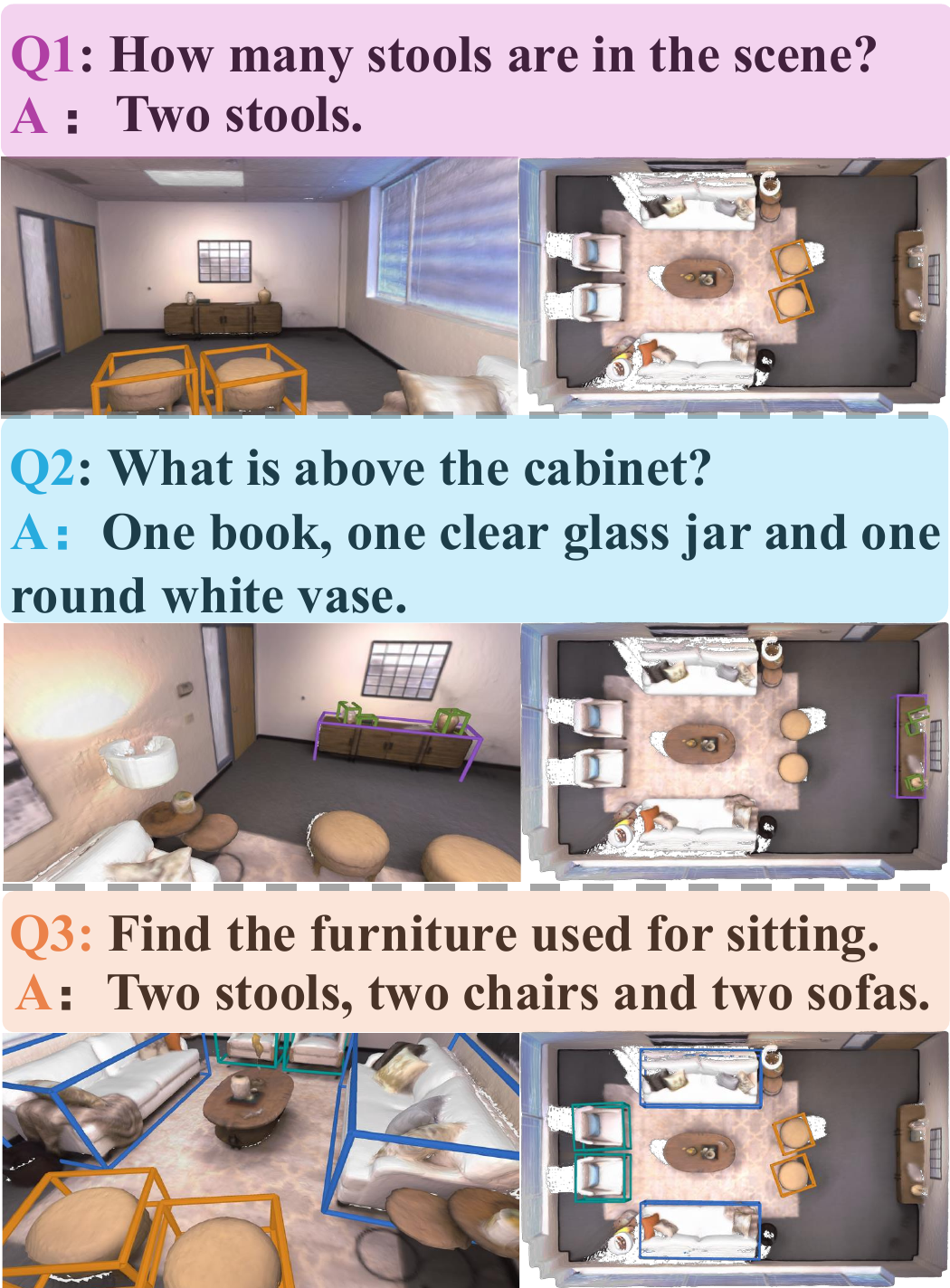}
    \caption{Scene understanding applications.}
\label{fig:applications}
\end{wrapfigure}
Our voxel semantic map provides a queryable 3D spatial representation
that supports both semantic reasoning and metric 3D grounding.
Figure~\ref{fig:applications} illustrates these complementary capabilities.

\noindent\textbf{Open-Vocabulary Scene Question Answering.}
The examples cover object counting (``How many stools are in the scene?''),
spatial relations (``What is above the cabinet?''), and functional
descriptions (``Find the furniture used for sitting.'').
The responses identify two stools, the objects above the cabinet, and
three categories of seating furniture, respectively. These examples
illustrate how the reconstructed semantic map supports language-driven
access to scene contents, including queries that span multiple object
categories and express relationships between objects.

\noindent\textbf{Language-Guided 3D Object Localization.}
Language features in the voxel map are associated with explicit 3D
positions, allowing queried concepts to be grounded in the reconstructed
geometry. The corresponding 3D boxes localize the retrieved objects in
perspective (left) and top-down (right) views, linking each answer to its spatial
support in a common scene coordinate system. Together, these examples
demonstrate the utility of open-vocabulary mapping for scene understanding
and spatial object retrieval beyond segmentation evaluation.

\noindent\textbf{Implementation.}
Object queries are encoded using the same language encoder as the semantic
map and matched against entity-level voxel embeddings by semantic similarity.
For counting queries, we count the retrieved entity instances. Spatial queries
are answered using the 3D locations and extents of the retrieved entities.
For functional descriptions, a lightweight rule-based parser maps the query
to a small set of candidate object concepts, which are then retrieved from the
open-vocabulary voxel map. The spatial extent of the retrieved voxels is used
to construct the 3D bounding boxes shown in both views.

\subsection{Limitations}

Our current system assumes a largely static scene and does not explicitly
model independently moving objects. Extending the framework to dynamic
environments remains an important direction for future work.

Our method also relies on foundation-model priors for geometry and semantics. Although confidence-aware fusion and consistency checks are introduced to
reduce the influence of unreliable predictions, the overall performance is
still bounded by the capability and generalization of the underlying
foundation models.

\finishwrap
\subsection{ADDITIONAL METHOD DETAILS}
\label{app:method_details}
\subsubsection{Geometry-Enhanced Monocular Gaussian SLAM}
\label{app:enhanced_geometry}
This section expands the geometry enhancement in
Section~\ref{sec:geometry}. We use camera-to-world poses
$T_t=[R_t,\mathbf t_t]$, with superscripts $S$ and $F$ denoting local
SLAM estimates and foundation predictions, respectively. All depth and
pose alignment is relative to the current SLAM coordinate system.

\paragraph{Context and anchor selection.}
We maintain a bounded foundation context consisting of recent keyframes
and geometrically complementary historical views, extending inference
beyond the local tracking window. Historical views are selected according
to confidence, geometric diversity under Eq.~\ref{eq:context_distance},
and overlap with the current context. A reliable overlapping historical
view serves as the alignment anchor, with reliability updated from
cross-window pose consistency. Overlapping observations align successive
predictions to a common coordinate system.

\paragraph{Depth alignment and gating.}
We align foundation depth to the current SLAM scale over reliable pixels
$\Omega_t^S\cap\Omega_t^F$ where both estimates are valid, using a
confidence-weighted robust fit of their log-depth difference:
\begin{equation}
\ell_t^*=\arg\min_\ell
\sum_{\mathbf u\in\Omega_t^S\cap\Omega_t^F} C_t^F(\mathbf u)
\rho_\delta\!\left(
\log D_t^S(\mathbf u)-\log D_t^F(\mathbf u)-\ell\right),
\qquad
\widetilde D_t^F=e^{\ell_t}D_t^F,
\label{eq:app_scalar_depth_alignment}
\end{equation}
where $\rho_\delta$ is the Huber penalty and $\ell_t$ is the temporally
smoothed scale estimate. For jointly valid pixels, define
$e_t=|\log\widetilde D_t^F-\log D_t^S|$ and
$\bar C_t^F=\operatorname{clip}((C_t^F-c_l)/(c_h-c_l),0,1)$,
with confidence thresholds $c_l<c_h$. The two-source confidence-and-consistency gate is
\begin{equation}
\alpha_t(\mathbf u)=
\begin{cases}
0, & C_t^F<c_l \text{ or } e_t>\tau_e,\\
1, & C_t^F\geq c_h\text{ and }e_t\leq\tau_e,\\
\bar C_t^F\exp(-e_t/\tau_e), & \text{otherwise}.
\end{cases}
\label{eq:app_complete_depth_gate}
\end{equation}
The gate first produces the blended depth $\bar D_t$ in Eq.~\ref{eq:depth_fusion}. We represent $s_t$ as a per-frame
low-resolution scale field with bilinear parameterization in log space,
fitted to reliable SLAM-to-blended-depth ratios and temporally smoothed.
Applying $s_t^{\alpha_t}=\exp(\alpha_t\log s_t)$ modulates the correction
in log-scale space and preserves valid SLAM depth wherever the prior is
rejected. For jointly valid pixels, the resulting mapping depth is
$D_t^M(\mathbf u)=s_t(\mathbf u)^{\alpha_t(\mathbf u)}\bar D_t(\mathbf u)$.
The disagreement gate applies only when both depth sources are valid. When local SLAM depth is missing, a valid aligned foundation prior is used with its confidence-modulated scale correction; when the foundation prior is unavailable or rejected, valid local SLAM depth is retained unchanged. Pixels with neither valid source are excluded from mapping supervision. The final depth $D_t^M$ initializes and supervises Gaussian mapping.

\paragraph{Long-horizon pose guidance.}
Historical context supplies trajectory proposals beyond the local tracking
window. We transfer these proposals to the SLAM frame through a
$\mathrm{Sim}(3)$ alignment over common frames $\mathcal I$:
\begin{equation}
(s_a,R_a,\mathbf t_a)=\arg\min_{s>0,R\in\mathrm{SO}(3),\mathbf t}
\sum_{i\in\mathcal I}
\|\mathbf t_i^S-(sR\mathbf t_i^F+\mathbf t)\|_2^2.
\label{eq:app_pose_alignment}
\end{equation}
The aligned poses are
$\widetilde{\mathbf t}_i^F=s_aR_a\mathbf t_i^F+\mathbf t_a$ and
$\widetilde R_i^F=R_aR_i^F$. Acceptance depends on trajectory coverage,
continuity, translation and rotation agreement, and scale stability,
using estimated geometry without ground-truth poses. Let
$\mathcal A\in\{0,1\}$ denote acceptance and
$\gamma=\mathcal A\gamma_0$ the guidance coefficient. We initialize
global visual bundle adjustment with
\begin{equation}
\begin{aligned}
\mathbf t_i^0&=(1-\gamma)\mathbf t_i^S
+\gamma\widetilde{\mathbf t}_i^F,\\
R_i^0&=R_i^S\operatorname{Exp}\!\left[
\gamma\operatorname{Log}\!\left((R_i^S)^\top\widetilde R_i^F\right)
\right].
\end{aligned}
\label{eq:app_pose_initialization}
\end{equation}
Accepted proposals provide long-horizon guidance to the initialization;
subsequent optimization refines poses and depths through visual
correspondence constraints, allowing observation-driven correction of
residual foundation-prediction errors. Rejected proposals retain the
SLAM initialization.

\paragraph{Gaussian map optimization.}
We build on the base Gaussian mapper~\citep{sandstrom2025splat} and initialize new
primitives from the fused mapping depth $D_t^M$. Gaussian parameters are
optimized using photometric, depth, and normal supervision, with standard
densification and pruning. Following backend pose updates, primitives are
re-anchored to their spawning keyframes and globally refined. Unless
otherwise stated, Gaussian optimization follows the base mapper; our
semantic-guided refinement is applied subsequently.

\subsubsection{Spatially Decoupled Open-Vocabulary Semantic Mapping}
\label{app:semantic_mapping_details}
We use Gaussian-rendered depth for semantic lifting so that region
observations are attached to the current reconstructed surface.
Foundation geometry improves lifting indirectly through the fused depth
supervision of Gaussian geometry and the accepted pose guidance.

The semantic map retains observation provenance so that accumulated
language evidence can follow changes in reconstructed geometry. Each
region observation $o=(t,r)$ stores its original mask $M_o$, normalized
language embedding $\mathbf e_o$, confidence $c_o$, and its current voxel
contributions. The region-language frontend follows
\citet{deng2025omnimap}, with $c_o$ combining detector and mask confidence.

\paragraph{Projection.}
Processed region masks are lifted using a consistent snapshot of
Gaussian-rendered depth, pose, and intrinsics. Valid pixels are
back-projected as $\mathbf x=R_t(\widehat D_tK_t^{-1}\widetilde{\mathbf u})
+\mathbf t_t$ and assigned to voxel keys $\lfloor\mathbf x/h\rfloor$.
The resulting support $\mathcal Q_o$ contains unique keys, so each region
contributes at most once per voxel. Mask boundaries and invalid depths
are excluded before projection.

\paragraph{Weighting.}
We use $w_{ov}=\max(c_o,\epsilon)\,\mathbf{1}\!\left\{v\in\mathcal Q_o\right\}$. and accumulate $\mathbf z_v=\sum_o w_{ov}\mathbf e_o$ and $W_v=\sum_o w_{ov}$.
Open-vocabulary retrieval compares normalized $\mathbf z_v$ with text
embeddings. We retain the unnormalized statistics and their contributing
observation identifiers, allowing each observation's contribution to be
retracted independently of other evidence in the voxel.

\paragraph{Entity association.}
For an observation $o$, we consider entities supported in its spatial
neighborhood. With entity support $\mathcal Q_e$ and voxel neighborhood
$\mathcal N(v)$, association combines spatial overlap and language
similarity:
\begin{equation}
\begin{aligned}
g(o,e)
&=
\frac{1}{|\mathcal Q_o|}
\sum_{v\in\mathcal Q_o}
\mathbf{1}\!\left\{
\mathcal N(v)\cap\mathcal Q_e\neq\emptyset
\right\},\\
a(o,e)
&=
\lambda_g g(o,e)
+
\lambda_f\,\mathbf e_o^\top\mathbf f_e .
\end{aligned}
\label{eq:app_entity_association}
\end{equation}
where $\mathbf f_e$ is the entity's normalized aggregated embedding.
The highest-scoring candidate is accepted above a threshold; otherwise,
a new entity is created. Entity identities organize object-level support
and constrain compatible surface neighborhoods; open-vocabulary retrieval
operates directly on voxel embeddings.

\paragraph{Observation revision and update scheduling.}
For each integrated frame, we retain the original semantic packet, its
observation identifiers and voxel supports, and the geometry snapshot used
at fusion. Semantic revision is triggered for observations whose
associated pose, depth, or calibration changes exceed predefined
thresholds; revisions are processed asynchronously to preserve online
operation.

Revision operates on the stored observations. We first retract each
affected observation's old contribution, then reproject its cached mask
with the updated geometry and insert the new contribution. For an
observation $o$, the feature update follows
Eq.~\ref{eq:semantic_revision}, while its weight is updated as
\begin{equation}
W_v'=W_v-w_{ov}^{\mathrm{old}}+w_{ov}^{\mathrm{new}}.
\label{eq:app_semantic_weight_revision}
\end{equation}
Observation membership, confidence statistics, and entity votes are
updated together, and voxels with no remaining support are removed.
The observation identifier is preserved: reprojection relocates existing
evidence without counting it as a new observation. Cached masks and
embeddings are reused, and unaffected observations retain their existing
contributions. Retraction from unnormalized accumulators preserves other
observations' evidence even when their supports overlap.

This incremental mechanism maintains semantic evidence during ongoing
tracking and mapping, including revisions of historical views, without
re-encoding their semantics. After standard Gaussian refinement, we
wait for pending asynchronous semantic updates to complete and rebuild
the voxel map from cached observations using the refined camera poses
and Gaussian-rendered depths. Semantic-guided refinement then operates
on the rebuilt map, followed by a final voxel-map rebuild to reflect
the updated geometry.

\subsubsection{Semantic-Guided Gaussian Geometry Refinement}
\label{app:semantic_geometry_details}
We detail the depth-reference selection and validity checks used for
the geometric evidence in Section~\ref{sec:gaussian_mapping}.
Reference-depth back-projection constructs local surface targets for
refinement; it does not insert new observations into the semantic voxel
map.

\paragraph{Depth-reference selection.}
At the start of each refinement stage, we snapshot the available depth
evidence and its camera geometry. When predicted mapping depth is
available, the reference is $D_t^E=D_t^M$, using the aligned and gated
depth from Appendix~\ref{app:enhanced_geometry}. We retain the associated confidence estimates and track the provenance of each depth reference, excluding predictions with unknown provenance. When predicted depth is unavailable, we instead use a frozen depth rendering of the current Gaussian map. Such a reference can
support local surface self-consistency, but cannot supply free-space
counter-evidence for floater removal. It introduces no independent depth
measurement. The reference depths, reliability masks, and derived surface
targets remain fixed during the corresponding optimization stage.

\paragraph{Camera and pixel validity.}
We ensure that depth references and semantic observations are spatially registered under consistent camera intrinsics, extrinsics, and image coordinates. Intrinsics are adjusted
consistently under image resizing, and inconsistent camera/depth snapshots
are excluded.

\paragraph{Depth-reference reliability.}
We retain finite positive depths with reliable confidence and suppress
evidence near depth discontinuities and holes. Predicted references use
their associated depth confidence, while rendered references additionally
require sufficient opacity coverage. Depth agreement uses the tolerance
$\delta_t(\mathbf u)=\max(\tau_d h,\tau_{\mathrm{rel}}D_t^E(\mathbf u))$,
combining voxel-scale and depth-relative uncertainty.

\paragraph{Validity for surface estimation.}
We project each candidate semantic voxel into the reference views and
sample the nearest image pixel. A view contributes a surface sample only
if the projection lies inside the image, its camera depth agrees with
$D_t^E$ within $\delta_t$, and both depth and semantic weights are
positive. Semantic compatibility requires
$\mathbf f_v^\top\mathbf e_o\geq\tau_{\mathrm{sem}}$, where
$\mathbf f_v$ and $\mathbf e_o$ are the normalized voxel and region
embeddings, respectively. For a voxel with an established entity
identity, the region must additionally be associated with the same
entity. We back-project the sampled reference depth to recover a 3D surface point, while using the voxel center only to identify the corresponding image ray. Samples are weighted by depth reliability and
semantic confidence. Their weighted spatial variance must remain below
a voxel-scaled threshold, and sufficient distinct views must support
them before they enter the local plane fit. The subsequent PCA rejects
insufficient, nearly collinear, or nonplanar neighborhoods, as well as
planes too far from the Gaussian being constrained.

\paragraph{Validity for free-space counter-evidence.}
Weak semantic support in sparse Gaussian neighborhoods proposes floater
candidates; removal requires geometric or photometric counter-evidence.
Only reliable predicted mapping depths provide free-space evidence.
We conservatively test the projected spatial extent of each Gaussian
against the reference depth rather than using its center alone.
Let $z_{tj}$ denote its center's camera depth and
$\sigma_{tj}^z=\sqrt{\mathbf a_t^\top\Sigma_j\mathbf a_t}$ its extent
standard deviation along the camera depth axis $\mathbf a_t$ in world
coordinates. A free-space contradiction requires the finite extent to
lie in front of the reliable reference surface:
\begin{equation}
z_{tj}+2\sigma_{tj}^z
<D_t^E(\mathbf u)-\delta_t(\mathbf u).
\label{eq:app_refinement_free_space}
\end{equation}
The contradiction must recur across geometrically distinct views.
Gaussians behind the reference surface may be occluded and provide no
free-space counter-evidence; invalid depth or missing semantic coverage
also cannot establish free space.

\paragraph{Photometric verification.}
For photometric verification, we evaluate the gradient of the RGB
reconstruction error with respect to each candidate's opacity logit over
views in which it contributes to the rendering. Consistently positive
gradients across sufficiently many distinct views, together with a
positive aggregate gradient, provide evidence that reducing its opacity
would decrease the reconstruction error. This is local directional
evidence; pruning additionally requires sufficiently low opacity and
renewed multi-view counter-evidence.
\subsection{ADDITIONAL EXPERIMENTAL DETAILS}
\label{app:experiment_details}

\subsubsection{Dataset Details}
\label{app:dataset_details}
\noindent\textbf{Replica.}
Replica~\citep{straub2019replica} provides digital reconstructions of
indoor spaces. We evaluate on eight scenes: \texttt{office0},
\texttt{office1}, \texttt{office2}, \texttt{office3}, \texttt{office4},
\texttt{room0}, \texttt{room1}, and \texttt{room2}, comprising five
office scenes and three room scenes.

\noindent\textbf{ScanNet.}
ScanNet~\citep{dai2017scannet} contains annotated real-world indoor scans.
Following the scene selection of Open-set SLAM~\citep{yan2026opensetslam},
we evaluate on six scenes: \texttt{scene0000\_00},
\texttt{scene0059\_00}, \texttt{scene0106\_00}, \texttt{scene0169\_00},
\texttt{scene0181\_00}, and \texttt{scene0207\_00}.

\finishwrap

\begin{wraptable}{r}{0.46\textwidth}
\vspace{0pt}
\centering
\caption{KITTI-360 segments with inclusive source-frame ranges.}
\label{tab:app_kitti360_segments}
\par\vspace{6pt}
\setlength{\tabcolsep}{3pt}
\renewcommand{\arraystretch}{1.08}
\tablefont
\begin{tabular*}{\linewidth}{@{\extracolsep{\fill}}lrr@{}}
\toprule
Sequence & Frame range & Frames \\
\midrule
\texttt{0002} & 13419--13718 & 300 \\
\texttt{0004} & 1946--2245 & 300 \\
\texttt{0004} & 6192--6491 & 300 \\
\texttt{0005} & 2030--2329 & 300 \\
\bottomrule
\end{tabular*}
\par
\normalsize
\end{wraptable}
\noindent\textbf{KITTI-360.}
KITTI-360~\citep{liao2022kitti360} is an outdoor urban-driving dataset
with long trajectories spanning diverse street environments.
Our system assumes a static scene and does not explicitly handle dynamic
objects. We therefore select continuous segments without moving
pedestrians or vehicles, yielding four evaluation scenes from three
original driving sequences. Each scene contains 300 consecutive
rectified RGB frames from camera \texttt{image\_00}, with no intermediate
frames skipped. Table~\ref{tab:app_kitti360_segments} lists the original
sequence identifiers and inclusive source-frame ranges; the two segments
from sequence \texttt{0004} are evaluated separately.
\finishwrap

\subsubsection{Evaluation Details}
\label{app:evaluation_details}
\noindent\textbf{Closed-set semantic evaluation.}
Following Open-set SLAM~\citep{yan2026opensetslam}, we evaluate on four
Replica scenes (room0, room1, room2, and office0) with ground-truth
semantic supervision. Ground-truth semantic labels replace predictions from the semantic frontend, while the downstream Gaussian and voxel mapping procedures remain unchanged. We report 2D segmentation mIoU in percent.
For its closed-set evaluation, Open-set SLAM trains a classification
head on rendered semantic features using cross-entropy supervision
from ground-truth labels.
We use the results reported by Open-set SLAM~\citep{yan2026opensetslam}
for Open-set SLAM itself and for the compared baselines
SGS-SLAM~\citep{li2024sgsslam} and GS3LAM~\citep{li2024gs3lam}.
We reproduce SemGauss-SLAM~\citep{zhu2025semgauss} and
FreeOcc~\citep{jiang2026freeocc} using their publicly available code
with ground-truth semantic supervision.
For OpenGS-SLAM~\citep{yang2025opengsslam}, we use the
\emph{Ours (Prior)} results reported in its Table~I, where all methods
use ground-truth semantic labels as semantic priors for mapping.

\noindent\textbf{Open-set semantic evaluation.}
To match our input setting, we restrict the quantitative comparison to
systems operating with RGB-only input and no externally supplied camera
poses. We consider OpenGS-SLAM~\citep{yang2025opengsslam},
Open-set SLAM~\citep{yan2026opensetslam},
RADIO-ViPE~\citep{nasser2026radiovipe}, and FreeOcc~\citep{jiang2026freeocc}
as related open-vocabulary semantic mapping systems.
At the time of our experiments, complete SLAM implementations and
evaluation code for OpenGS-SLAM and Open-set SLAM were not publicly
available, preventing reproducible comparisons under a common protocol.
We therefore compare against RADIO-ViPE and FreeOcc using their
publicly available implementations in RGB-only configurations, with
camera poses estimated by each system.
We adopt the 3D semantic segmentation protocol used by RADIO-ViPE,
which follows RayFronts~\citep{alama2025rayfronts}, and apply it to
Replica and ScanNet. Following the without-background setting in
RADIO-ViPE, we exclude background categories, including structural
classes such as walls and floors, from evaluation.
We report mean class accuracy (mAcc), overall accuracy (fAcc), mean
IoU (mIoU), and frequency-weighted IoU (fIoU), all on a 0--1 scale.

\noindent\textbf{Time Efficiency evaluation.}
We evaluate the end-to-end runtime of our method and four publicly
available baselines: SNI-SLAM~\citep{zhu2024snislam},
SGS-SLAM~\citep{li2024sgsslam}, SemGauss-SLAM~\citep{zhu2025semgauss},
and GS3LAM~\citep{li2024gs3lam}.
All five methods are measured on a single NVIDIA RTX A6000 GPU under
the same experimental setup and with a common end-to-end timing scope.
Timing includes data processing, semantic segmentation, tracking and
mapping, and the final refinement stages used to produce the evaluated
reconstruction. Each baseline retains the iteration counts and parameter
settings specified in its paper and released implementation.

Open-set SLAM~\citep{yan2026opensetslam} is reported as a
literature-derived estimate because its complete SLAM implementation was
not publicly available at the time of our experiments.
Its approximately 90-minute runtime is obtained from
$T=N/(60\,\mathrm{FPS})$, using $N=2000$ frames and the reported
$0.37$ FPS for the LoopSplat-based variant on Replica room0.
The original measurement uses a single NVIDIA RTX 3090 GPU.

\subsubsection{Detailed Reconstruction Results}
\label{app:detailed_reconstruction}
We provide per-scene rendering quality (PSNR, SSIM, and LPIPS) and
geometric reconstruction error (L1 depth) on Replica and ScanNet,
as well as per-scene rendering quality on KITTI-360.
Tables~\ref{tab:app_replica_render_nkf} and~\ref{tab:app_depth_nkf}
report the Replica results, while
Tables~\ref{tab:app_scannet_render_nkf} and~\ref{tab:app_scannet_depth_nkf}
report the ScanNet results. KITTI-360 rendering results are presented
in Table~\ref{tab:app_kitti360_render_kf}.

We further provide additional qualitative results.
Figure~\ref{fig:app_mesh} compares reconstructed meshes with those
of baseline methods and the ground truth.
Figure~\ref{fig:app_semantic} visualizes our closed-set Gaussian
semantics and open-set voxel semantics alongside RGB images,
depth maps, and ground-truth semantic annotations.
Figure~\ref{fig:app_rgb} presents RGB rendering comparisons on
ScanNet and KITTI-360, with enlarged regions highlighting local details.

\subsubsection{More Ablation Studies}
\noindent\textbf{Overall Component Ablation.}
We evaluate the contributions of geometry enhancement and semantic-guided
geometry refinement on all eight Replica scenes.
Our SLAM backbone follows the architecture of DROID-SLAM~\citep{teed2021droid}
and related Gaussian SLAM systems~\citep{homeyer2024droidsplat,zhang2025hi}.
Table~\ref{tab:ablation_components} reports scene-averaged results as we
progressively add the two components to this backbone. The first row
disables both components, and the last row includes both.

Adding geometry enhancement increases PSNR from 36.79 to 38.37\,dB
($+1.58$\,dB) and SSIM from 0.958 to 0.974, while reducing LPIPS
from 0.036 to 0.030. It also reduces L1 depth error from 1.564 to
1.260\,cm, a 19.4\% reduction.
Adding semantic-guided refinement further improves PSNR to 38.65\,dB
($+0.28$\,dB) and SSIM to 0.975, and lowers LPIPS to 0.026.
L1 depth error decreases further to 1.170\,cm, a 7.1\% reduction
relative to geometry enhancement alone.
These results show that geometry enhancement improves both appearance
and depth reconstruction, while semantic-guided refinement provides
additional gains across all four metrics.

\noindent\textbf{Effect of Semantic Guidance.}
To isolate the contribution of semantic guidance, we compare our full
refinement with a geometry-only variant that retains the same geometric
refinement procedure but removes semantic cues for candidate and
neighborhood selection.
As shown in Table~\ref{tab:semantic_guidance_refinement},
geometry-only refinement yields a modest improvement over no refinement,
whereas semantic guidance further reduces Acc.\ from 12.07 to 9.79\,cm
and improves F-score from 49.14\% to 50.97\%.
This indicates that the gain is not merely due to additional geometric
refinement; semantic cues improve the selection of reliable floater
candidates and compatible surface neighborhoods.

\begin{table}[t]
\centering
\caption{Progressive ablation of geometry enhancement and semantic-guided
refinement, averaged over all eight Replica scenes.}
\label{tab:ablation_components}
\setlength{\tabcolsep}{6.5pt}
\renewcommand{\arraystretch}{1.12}
\begin{tabular}{l|cccc}
\toprule
Method
& PSNR [dB] $\uparrow$
& SSIM $\uparrow$
& LPIPS $\downarrow$
& L1 Depth [cm] $\downarrow$ \\
\midrule

w/o both
& 36.79
& 0.958
& 0.036
& 1.564 \\

+ Geometry Enhancement
& 38.37
& 0.974
& 0.030
& 1.260 \\

+ Semantic-Guided Refinement
& \textbf{38.65}
& \textbf{0.975}
& \textbf{0.026}
& \textbf{1.170} \\

\bottomrule
\end{tabular}
\end{table}

\begin{table}[t]
\centering
\caption{
Ablation of semantic guidance in Gaussian geometry refinement on
ScanNet scene0207. The geometry-only variant retains the same geometric
refinement while removing semantic guidance.
}
\label{tab:semantic_guidance_refinement}
\setlength{\tabcolsep}{6pt}
\renewcommand{\arraystretch}{1.05}
\begin{tabular}{lcc}
\toprule
\textbf{Refinement Variant}
& \textbf{Acc. [cm] $\downarrow$}
& \textbf{F-score [\%] $\uparrow$} \\
\midrule
No Refinement
& 13.76
& 48.86 \\
Geometry-Only Refinement
& 12.07
& 49.14 \\
Semantic-Guided Refinement (ours)
& \textbf{9.79}
& \textbf{50.97} \\
\bottomrule
\end{tabular}
\end{table}

\subsection{LEGAL AND ETHICAL CONSIDERATIONS}
\label{app:legal_ethical}
\subsubsection{Asset License}
Replica and ScanNet are available under the
\href{https://github.com/facebookresearch/Replica-Dataset/blob/main/LICENSE}{Replica Dataset Research Terms}
and \href{https://kaldir.vc.cit.tum.de/scannet/ScanNet_TOS.pdf}{ScanNet Terms of Use}, respectively.
\href{https://www.cvlibs.net/datasets/kitti-360/}{KITTI-360}
is released under CC BY-NC-SA 3.0.
For pretrained models,
\href{https://huggingface.co/depth-anything/DA3NESTED-GIANT-LARGE}{Depth Anything 3 (DA3NESTED-GIANT-LARGE)}
uses CC BY-NC 4.0,
\href{https://github.com/AILab-CVC/YOLO-World}{YOLO-World} uses GPL-3.0,
and \href{https://github.com/baaivision/tokenize-anything}{Tokenize Anything}
and \href{https://huggingface.co/sentence-transformers/all-MiniLM-L6-v2}{Sentence-BERT (all-MiniLM-L6-v2)}
use Apache-2.0.

\clearpage
\begingroup
\setlength{\floatsep}{12pt}
\setlength{\textfloatsep}{12pt}
\setlength{\intextsep}{12pt}
\makeatletter
\setlength{\@fptop}{0pt}
\setlength{\@fpsep}{12pt}
\setlength{\@fpbot}{0pt plus 1fil}
\makeatother
\begin{table*}[!htbp]
\centering
\caption{Rendering quality on Replica.}
\label{tab:app_replica_render_nkf}
\setlength{\tabcolsep}{4.2pt}
\renewcommand{\arraystretch}{0.88}
\resizebox{\textwidth}{!}{%
\begin{tabular}{llccccccccc}
\toprule
\textbf{Method} & \textbf{Metric} &
\textbf{ro-0} & \textbf{ro-1} & \textbf{ro-2} &
\textbf{of-0} & \textbf{of-1} & \textbf{of-2} &
\textbf{of-3} & \textbf{of-4} & \textbf{Avg.} \\
\midrule

\multirow{3}{*}{Splat-SLAM}
& PSNR $\uparrow$
& \cellcolor{third}30.75
& 32.69
& \cellcolor{third}32.63
& \cellcolor{third}39.00
& 36.15
& \cellcolor{third}33.69
& 33.52
& 34.73
& 34.15 \\
& SSIM $\uparrow$
& \cellcolor{third}0.914
& 0.930
& \cellcolor{third}0.944
& \cellcolor{second}0.973
& 0.964
& 0.930
& 0.943
& \cellcolor{third}0.959
& \cellcolor{third}0.945 \\
& LPIPS $\downarrow$
& \cellcolor{third}0.091
& 0.076
& \cellcolor{third}0.070
& \cellcolor{third}0.046
& 0.066
& 0.113
& 0.066
& \cellcolor{third}0.066
& 0.074 \\
\midrule

\multirow{3}{*}{ARTDECO}
& PSNR $\uparrow$
& 29.28 & 28.34 & 30.44 & 35.63 & 34.38 & 27.88 & 28.29 & 28.40 & 30.33 \\
& SSIM $\uparrow$
& 0.883 & 0.875 & 0.920 & 0.951 & 0.936 & 0.915 & 0.901 & 0.902 & 0.910 \\
& LPIPS $\downarrow$
& 0.172 & 0.182 & 0.153 & 0.137 & 0.189 & 0.188 & 0.173 & 0.174 & 0.171 \\
\midrule

\multirow{3}{*}{S3PO-GS}
& PSNR $\uparrow$
& 27.64 & 27.03 & 28.41 & 34.05 & 33.13 & 28.97 & 30.84 & 30.36 & 30.05 \\
& SSIM $\uparrow$
& 0.852 & 0.829 & 0.882 & 0.931 & 0.935 & 0.902 & 0.914 & 0.918 & 0.895 \\
& LPIPS $\downarrow$
& 0.196 & 0.266 & 0.220 & 0.171 & 0.195 & 0.224 & 0.163 & 0.191 & 0.203 \\
\midrule

\multirow{3}{*}{DROID-Splat}
& PSNR $\uparrow$
& 30.68
& \cellcolor{third}34.36
& 32.46
& 37.80
& \cellcolor{third}38.93
& 33.34
& \cellcolor{third}33.79
& \cellcolor{third}35.08
& \cellcolor{third}34.56 \\
& SSIM $\uparrow$
& 0.885
& \cellcolor{third}0.950
& 0.937
& 0.963
& \cellcolor{third}0.974
& \cellcolor{third}0.939
& \cellcolor{third}0.955
& 0.949
& 0.944 \\
& LPIPS $\downarrow$
& 0.130
& \cellcolor{third}0.056
& 0.089
& 0.057
& \cellcolor{third}0.045
& \cellcolor{third}0.080
& \cellcolor{third}0.061
& 0.067
& \cellcolor{third}0.073 \\
\midrule

\multirow{3}{*}{FreeOcc}
& PSNR $\uparrow$
& 15.89 & 19.20 & 17.99 & 24.48 & 24.34 & 15.91 & 16.43 & 18.00 & 19.03 \\
& SSIM $\uparrow$
& 0.594 & 0.677 & 0.679 & 0.747 & 0.785 & 0.668 & 0.654 & 0.738 & 0.693 \\
& LPIPS $\downarrow$
& 0.369 & 0.274 & 0.312 & 0.193 & 0.206 & 0.306 & 0.304 & 0.276 & 0.280 \\
\midrule

\multirow{3}{*}{HI-SLAM2}
& PSNR $\uparrow$
& \cellcolor{second}34.46
& \cellcolor{second}36.45
& \cellcolor{second}37.13
& \cellcolor{second}41.42
& \cellcolor{second}42.66
& \cellcolor{second}35.82
& \cellcolor{second}35.87
& \cellcolor{second}37.98
& \cellcolor{second}37.72 \\
& SSIM $\uparrow$
& \cellcolor{second}0.951
& \cellcolor{second}0.955
& \cellcolor{second}0.966
& \cellcolor{third}0.967
& \cellcolor{second}0.981
& \cellcolor{second}0.955
& \cellcolor{second}0.964
& \cellcolor{second}0.971
& \cellcolor{second}0.964 \\
& LPIPS $\downarrow$
& \cellcolor{second}0.041
& \cellcolor{second}0.039
& \cellcolor{second}0.033
& \cellcolor{second}0.021
& \cellcolor{second}0.019
& \cellcolor{second}0.041
& \cellcolor{second}0.035
& \cellcolor{second}0.041
& \cellcolor{second}0.034 \\
\midrule

\multirow{3}{*}{\textbf{Ours}}
& PSNR $\uparrow$
& \cellcolor{best}35.24
& \cellcolor{best}37.48
& \cellcolor{best}38.29
& \cellcolor{best}42.55
& \cellcolor{best}43.53
& \cellcolor{best}36.74
& \cellcolor{best}36.52
& \cellcolor{best}38.83
& \cellcolor{best}38.65 \\
& SSIM $\uparrow$
& \cellcolor{best}0.963
& \cellcolor{best}0.972
& \cellcolor{best}0.976
& \cellcolor{best}0.987
& \cellcolor{best}0.986
& \cellcolor{best}0.972
& \cellcolor{best}0.972
& \cellcolor{best}0.974
& \cellcolor{best}0.975 \\
& LPIPS $\downarrow$
& \cellcolor{best}0.034
& \cellcolor{best}0.029
& \cellcolor{best}0.025
& \cellcolor{best}0.014
& \cellcolor{best}0.015
& \cellcolor{best}0.030
& \cellcolor{best}0.027
& \cellcolor{best}0.032
& \cellcolor{best}0.026 \\
\bottomrule
\end{tabular}}
\end{table*}

\begin{table*}[!htbp]
\centering
\caption{Depth reconstruction error on Replica (L1 depth $\downarrow$ [cm]).}
\label{tab:app_depth_nkf}
\setlength{\tabcolsep}{4.8pt}
\renewcommand{\arraystretch}{0.88}
\resizebox{\textwidth}{!}{%
\begin{tabular}{lccccccccc}
\toprule
\textbf{Method} &
\textbf{ro-0} & \textbf{ro-1} & \textbf{ro-2} &
\textbf{of-0} & \textbf{of-1} & \textbf{of-2} &
\textbf{of-3} & \textbf{of-4} & \textbf{Avg.} \\
\midrule
Splat-SLAM
& \cellcolor{third}2.93 & \cellcolor{third}2.17
& \cellcolor{third}2.19 & 2.50
& \cellcolor{third}1.79 & \cellcolor{third}2.83
& \cellcolor{third}2.76 & \cellcolor{third}2.57
& \cellcolor{third}2.468 \\

ARTDECO
& 7.11 & 6.70 & 6.04 & 4.37
& 4.42 & 10.62 & 8.83 & 52.43 & 12.565 \\

S3PO-GS
& 9.84 & 12.02 & 7.94 & 5.62
& 20.92 & 10.99 & 12.66 & 12.21 & 11.523 \\

DROID-Splat
& 4.58 & 2.75 & 3.78
& \cellcolor{third}2.06 & 2.01
& 10.82 & 15.60 & 3.93 & 5.690 \\

FreeOcc
& 16.90
& 8.24
& 14.22
& 7.90
& 6.37
& 13.05
& 19.33
& 16.56
& 12.820 \\

HI-SLAM2
& \cellcolor{second}1.93 & \cellcolor{second}1.17
& \cellcolor{second}1.47 & \cellcolor{second}1.32
& \cellcolor{second}0.86 & \cellcolor{second}1.54
& \cellcolor{second}2.19 & \cellcolor{second}1.75
& \cellcolor{second}1.528 \\

\textbf{Ours}
& \cellcolor{best}1.35 & \cellcolor{best}0.98
& \cellcolor{best}1.16 & \cellcolor{best}1.00
& \cellcolor{best}0.67 & \cellcolor{best}1.19
& \cellcolor{best}1.76 & \cellcolor{best}1.26
& \cellcolor{best}1.170 \\
\bottomrule
\end{tabular}}
\end{table*}

\begin{figure}[!htbp]
  \centering
  \includegraphics[width=\textwidth]{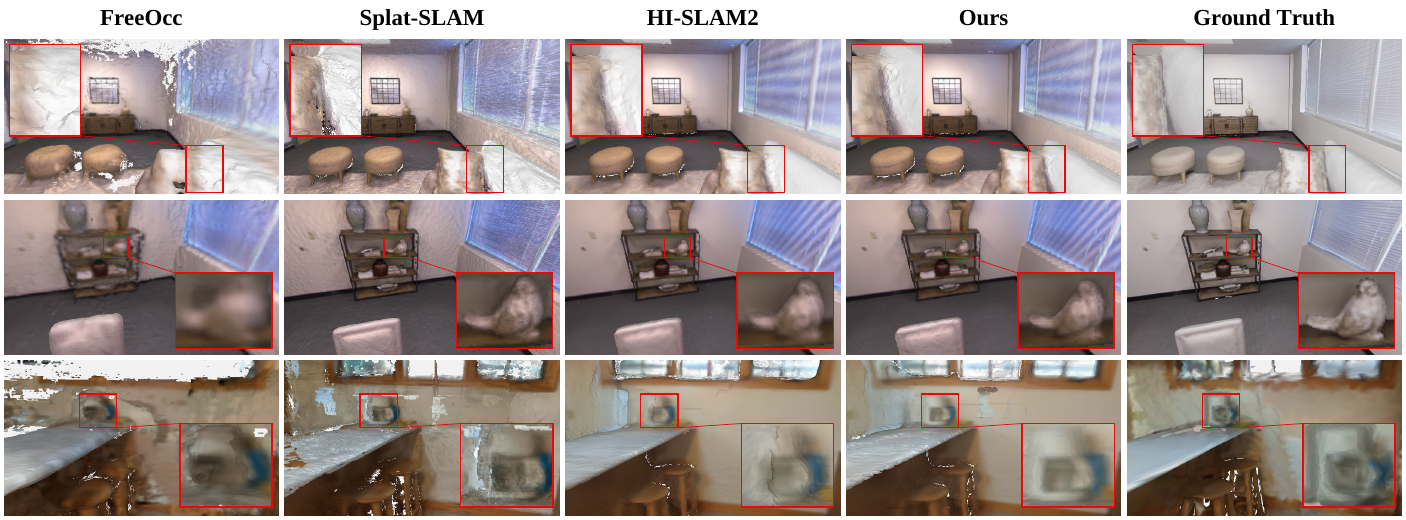}
  \caption{Additional qualitative comparisons of reconstructed
  meshes against baseline methods and the ground truth.}
  \label{fig:app_mesh}
\end{figure}

\clearpage
\begin{table*}[!htbp]
\centering
\caption{Rendering quality on ScanNet.}
\label{tab:app_scannet_render_nkf}
\setlength{\tabcolsep}{5.0pt}
\renewcommand{\arraystretch}{0.88}
\resizebox{\textwidth}{!}{%
\begin{tabular}{llccccccc}
\toprule
\textbf{Method} & \textbf{Metric} &
\textbf{0000} & \textbf{0059} & \textbf{0106} &
\textbf{0169} & \textbf{0181} & \textbf{0207} & \textbf{Avg.} \\
\midrule

\multirow{3}{*}{Splat-SLAM}
& PSNR $\uparrow$
& 23.29 & 21.30 & 22.49 & 24.66 & 17.93
& \cellcolor{third}22.72 & 22.07 \\
& SSIM $\uparrow$
& \cellcolor{third}0.800
& 0.793 & 0.803 & 0.801 & 0.762
& \cellcolor{third}0.794 & 0.792 \\
& LPIPS $\downarrow$
& \cellcolor{best}0.220
& \cellcolor{best}0.216
& \cellcolor{second}0.248
& \cellcolor{second}0.218
& \cellcolor{second}0.249
& \cellcolor{best}0.230
& \cellcolor{best}0.230 \\
\midrule

\multirow{3}{*}{ARTDECO}
& PSNR $\uparrow$
& 19.37 & 19.99 & 19.36 & 20.48 & 19.02 & 20.16 & 19.73 \\
& SSIM $\uparrow$
& 0.732
& \cellcolor{third}0.794
& 0.812
& \cellcolor{third}0.806
& 0.806 & 0.752 & 0.784 \\
& LPIPS $\downarrow$
& 0.363 & 0.343 & 0.341 & 0.324 & 0.397 & 0.372 & 0.357 \\
\midrule

\multirow{3}{*}{S3PO-GS}
& PSNR $\uparrow$
& 16.44 & 17.22 & 18.47 & 18.16
& \cellcolor{third}21.44 & 21.16 & 18.82 \\
& SSIM $\uparrow$
& 0.716 & 0.707 & 0.769 & 0.782
& \cellcolor{second}0.846
& \cellcolor{second}0.812
& 0.772 \\
& LPIPS $\downarrow$
& 0.689 & 0.566 & 0.530 & 0.575 & 0.505 & 0.563 & 0.571 \\
\midrule

\multirow{3}{*}{DROID-Splat}
& PSNR $\uparrow$
& \cellcolor{second}24.09
& \cellcolor{second}22.82
& \cellcolor{second}24.82
& \cellcolor{third}25.01
& \cellcolor{second}24.37
& \cellcolor{second}26.79
& \cellcolor{second}24.65 \\
& SSIM $\uparrow$
& 0.754 & 0.786
& \cellcolor{third}0.834
& 0.804 & \cellcolor{third}0.828 & 0.792
& \cellcolor{third}0.800 \\
& LPIPS $\downarrow$
& 0.321 & 0.289 & 0.259 & 0.300 & 0.331
& \cellcolor{second}0.267
& 0.295 \\
\midrule

\multirow{3}{*}{FreeOcc}
& PSNR $\uparrow$
& 15.90 & 12.65 & 11.55 & 15.86 & 14.72 & 15.46 & 14.36 \\
& SSIM $\uparrow$
& 0.459 & 0.365 & 0.359 & 0.517 & 0.511 & 0.466 & 0.446 \\
& LPIPS $\downarrow$
& 0.528 & 0.560 & 0.651 & 0.523 & 0.564 & 0.553 & 0.563 \\
\midrule

\multirow{3}{*}{HI-SLAM2}
& PSNR $\uparrow$
& \cellcolor{third}23.35
& \cellcolor{third}21.64
& \cellcolor{third}22.66
& \cellcolor{second}28.67
& 19.65 & 22.17
& \cellcolor{third}23.02 \\
& SSIM $\uparrow$
& \cellcolor{second}0.808
& \cellcolor{second}0.808
& \cellcolor{second}0.844
& \cellcolor{second}0.849
& \cellcolor{third}0.828
& 0.772
& \cellcolor{second}0.818 \\
& LPIPS $\downarrow$
& \cellcolor{third}0.271
& \cellcolor{third}0.265
& \cellcolor{third}0.250
& \cellcolor{third}0.225
& \cellcolor{third}0.273
& 0.327
& \cellcolor{third}0.269 \\
\midrule

\multirow{3}{*}{\textbf{Ours}}
& PSNR $\uparrow$
& \cellcolor{best}28.69
& \cellcolor{best}26.65
& \cellcolor{best}27.37
& \cellcolor{best}30.22
& \cellcolor{best}29.81
& \cellcolor{best}29.16
& \cellcolor{best}28.65 \\
& SSIM $\uparrow$
& \cellcolor{best}0.835
& \cellcolor{best}0.863
& \cellcolor{best}0.884
& \cellcolor{best}0.862
& \cellcolor{best}0.894
& \cellcolor{best}0.814
& \cellcolor{best}0.859 \\
& LPIPS $\downarrow$
& \cellcolor{second}0.254
& \cellcolor{second}0.228
& \cellcolor{best}0.214
& \cellcolor{best}0.208
& \cellcolor{best}0.230
& \cellcolor{third}0.302
& \cellcolor{second}0.239 \\
\bottomrule
\end{tabular}}
\end{table*}

\begin{table*}[!htbp]
\centering
\caption{Depth reconstruction error on ScanNet (L1 depth $\downarrow$ [cm]).}
\label{tab:app_scannet_depth_nkf}
\setlength{\tabcolsep}{6.0pt}
\renewcommand{\arraystretch}{0.88}
\resizebox{\textwidth}{!}{%
\begin{tabular}{lccccccc}
\toprule
\textbf{Method} &
\textbf{0000} & \textbf{0059} & \textbf{0106} &
\textbf{0169} & \textbf{0181} & \textbf{0207} & \textbf{Avg.} \\
\midrule

Splat-SLAM
& 8.89
& \cellcolor{third}12.57
& 24.53
& \cellcolor{second}12.36
& 16.83
& \cellcolor{third}12.80
& 14.66 \\

ARTDECO
& \cellcolor{third}8.45
& 15.67
& \cellcolor{third}18.72
& 15.33
& 12.71
& 15.92
& \cellcolor{third}14.47 \\

S3PO-GS
& 43.78
& 20.05
& 22.70
& 83.01
& 50.43
& 34.59
& 42.43 \\

DROID-Splat
& 17.59
& 17.33
& 19.00
& 31.13
& 12.65
& 18.70
& 19.40 \\

FreeOcc
& 12.66
& 26.35
& 31.87
& 19.87
& \cellcolor{third}11.23
& 16.31
& 19.72 \\

HI-SLAM2
& \cellcolor{best}5.24
& \cellcolor{second}10.16
& \cellcolor{second}14.92
& \cellcolor{third}13.83
& \cellcolor{second}6.90
& \cellcolor{second}11.75
& \cellcolor{second}10.47 \\

\textbf{Ours}
& \cellcolor{second}5.27
& \cellcolor{best}9.31
& \cellcolor{best}14.85
& \cellcolor{best}11.03
& \cellcolor{best}6.44
& \cellcolor{best}9.43
& \cellcolor{best}9.39 \\

\bottomrule
\end{tabular}}
\end{table*}

\clearpage
\begin{table*}[!htbp]
\centering
\caption{Rendering quality on KITTI-360.}
\label{tab:app_kitti360_render_kf}
\setlength{\tabcolsep}{5.0pt}
\renewcommand{\arraystretch}{0.88}
\resizebox{\textwidth}{!}{%
\begin{tabular}{llccccc}
\toprule
\textbf{Method} & \textbf{Metric} &
\textbf{scene01} & \textbf{scene02} & \textbf{scene03} &
\textbf{scene04} & \textbf{Avg.} \\
\midrule
\multirow{3}{*}{Splat-SLAM} & PSNR $\uparrow$ & \cellcolor{second}22.82 & 24.89 & 22.77 & \cellcolor{third}23.13 & 23.40 \\
 & SSIM $\uparrow$ & 0.740 & 0.744 & 0.683 & \cellcolor{third}0.748 & 0.729 \\
 & LPIPS $\downarrow$ & 0.331 & 0.353 & 0.430 & 0.353 & 0.367 \\
\midrule
\multirow{3}{*}{ARTDECO} & PSNR $\uparrow$ & 20.74 & 22.81 & 23.03 & 21.43 & 22.00 \\
 & SSIM $\uparrow$ & 0.698 & 0.705 & \cellcolor{third}0.720 & 0.731 & 0.714 \\
 & LPIPS $\downarrow$ & \cellcolor{second}0.268 & \cellcolor{second}0.284 & \cellcolor{best}0.276 & \cellcolor{best}0.285 & \cellcolor{second}0.278 \\
\midrule
\multirow{3}{*}{S3PO-GS} & PSNR $\uparrow$ & 19.58 & 22.20 & 20.19 & 20.61 & 20.65 \\
 & SSIM $\uparrow$ & 0.649 & 0.683 & 0.633 & 0.707 & 0.668 \\
 & LPIPS $\downarrow$ & 0.388 & 0.384 & 0.469 & 0.388 & 0.407 \\
\midrule
\multirow{3}{*}{DROID-Splat} & PSNR $\uparrow$ & 22.69 & \cellcolor{second}25.71 & \cellcolor{third}23.30 & \cellcolor{second}23.22 & \cellcolor{second}23.73 \\
 & SSIM $\uparrow$ & \cellcolor{third}0.759 & \cellcolor{third}0.768 & \cellcolor{third}0.720 & \cellcolor{second}0.760 & \cellcolor{third}0.752 \\
 & LPIPS $\downarrow$ & \cellcolor{third}0.305 & 0.301 & 0.362 & \cellcolor{third}0.327 & \cellcolor{third}0.324 \\
\midrule
\multirow{3}{*}{FreeOcc} & PSNR $\uparrow$ & 10.22 & 8.71 & 9.96 & 8.54 & 9.36 \\
 & SSIM $\uparrow$ & 0.370 & 0.286 & 0.296 & 0.353 & 0.326 \\
 & LPIPS $\downarrow$ & 0.471 & 0.562 & 0.541 & 0.559 & 0.533 \\
\midrule
\multirow{3}{*}{HI-SLAM2} & PSNR $\uparrow$ & \cellcolor{third}22.78 & \cellcolor{third}25.56 & \cellcolor{second}24.35 & 20.97 & \cellcolor{third}23.42 \\
 & SSIM $\uparrow$ & \cellcolor{second}0.771 & \cellcolor{second}0.778 & \cellcolor{second}0.749 & 0.719 & \cellcolor{second}0.754 \\
 & LPIPS $\downarrow$ & 0.309 & \cellcolor{third}0.299 & \cellcolor{third}0.334 & 0.412 & 0.339 \\
\midrule
\multirow{3}{*}{\textbf{Ours}} & PSNR $\uparrow$ & \cellcolor{best}24.70 & \cellcolor{best}27.64 & \cellcolor{best}25.39 & \cellcolor{best}23.84 & \cellcolor{best}25.39 \\
 & SSIM $\uparrow$ & \cellcolor{best}0.820 & \cellcolor{best}0.824 & \cellcolor{best}0.777 & \cellcolor{best}0.781 & \cellcolor{best}0.800 \\
 & LPIPS $\downarrow$ & \cellcolor{best}0.226 & \cellcolor{best}0.219 & \cellcolor{second}0.285 & \cellcolor{second}0.313 & \cellcolor{best}0.261 \\
\bottomrule
\end{tabular}}
\end{table*}

\begin{figure}[!htbp]
  \centering
  \includegraphics[width=\textwidth]{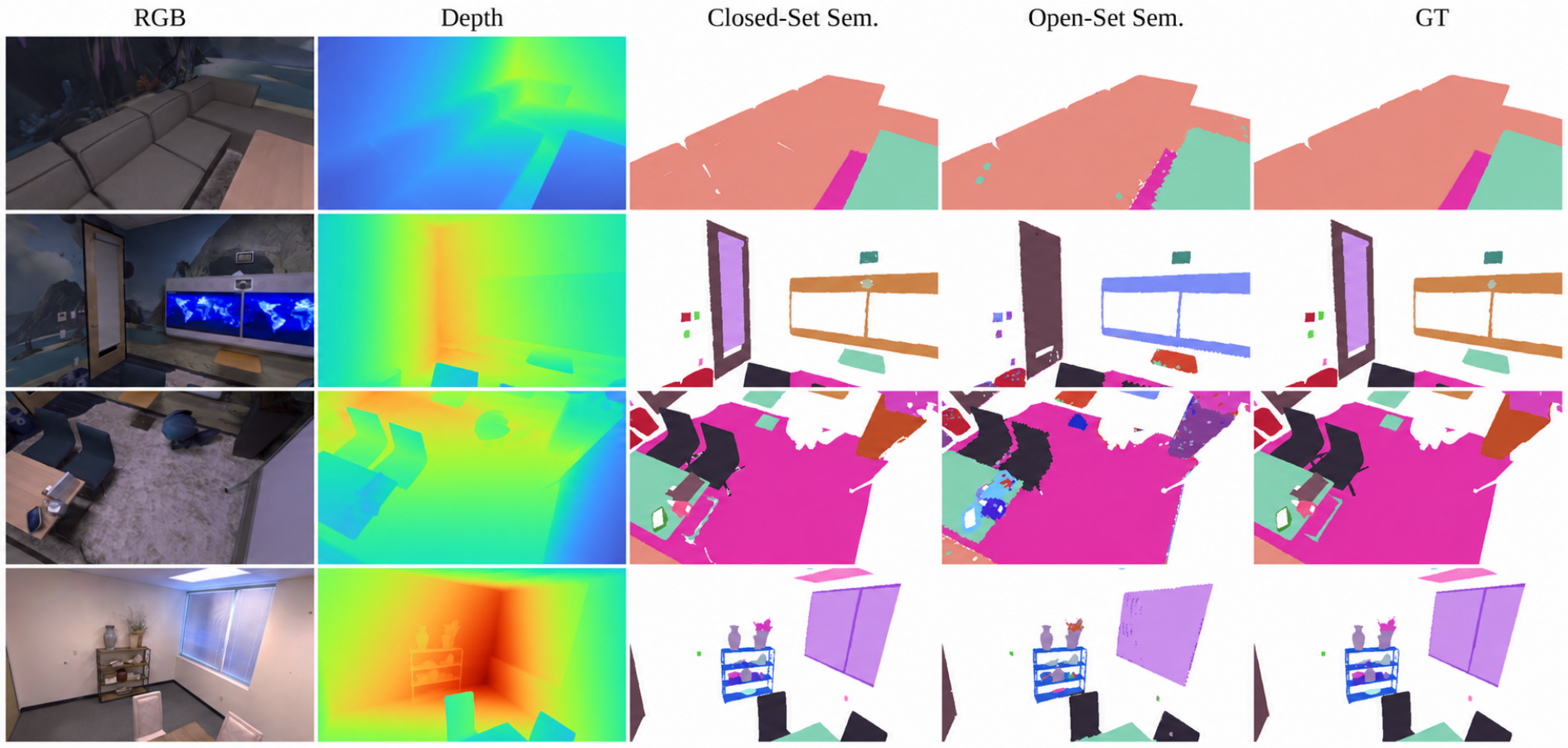}
  \caption{Additional semantic reconstruction results. From left
  to right: RGB images, depth maps, closed-set Gaussian semantics,
  open-set voxel semantics, and ground-truth semantic annotations.}
  \label{fig:app_semantic}
\end{figure}

\clearpage
\begin{figure}[p]
  \centering
  \includegraphics[width=\textwidth,height=0.88\textheight,
                   keepaspectratio]{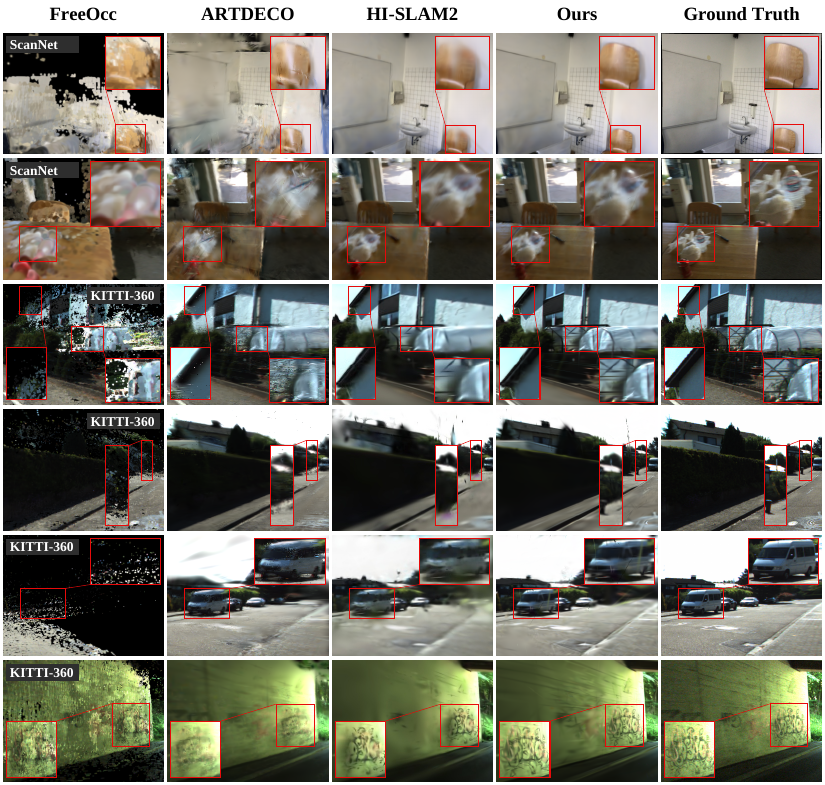}
  \caption{Additional RGB rendering comparisons on ScanNet and
  KITTI-360. Enlarged regions highlight local appearance details.}
  \label{fig:app_rgb}
\end{figure}
\clearpage
\endgroup

\end{document}